\documentclass{article}
\usepackage{ijcai26}

\usepackage{times}
\usepackage{soul}
\usepackage{url}
\usepackage[hidelinks]{hyperref}
\usepackage[utf8]{inputenc}
\usepackage[small]{caption}
\usepackage{graphicx}
\usepackage{amsmath}
\usepackage{amsthm}
\usepackage{booktabs}
\usepackage{algorithm}
\usepackage[switch]{lineno}
\usepackage{amssymb}
\usepackage{multirow} 
\usepackage{algpseudocode}
\usepackage{rotating}
\usepackage{makecell}
\usepackage{xcolor} 

\title{Advancing Edge Classification through High-Dimensional \\Causal Modeling of Node-Edge Interplay}

\author{
Duanyu Feng$^1$
\and
Li Ding$^1$\and
Hongru Liang$^1$\And
Wenqiang Lei$^1$\\
\affiliations
$^1$Sichuan University\\
\emails
\{fengduanyuscu, dinglee\}@stu.scu.edu.cn,
\{lianghongru, wenqianglei\}@scu.edu.cn
}

\begin{document}

\maketitle

\begin{abstract}
Edge classification, a crucial task for graph applications, remains relatively under-explored compared to link prediction. 
Current methods often overlook the potential causal influences of node features on edge features, leading to a loss of relevant prior information. 
In this work, we present an empirical exploration using the Causal Edge Classification Framework (CECF). 
Unlike conventional causal inference methods, CECF is the first framework to apply causal inference principles to the edge classification task and to explore modeling edge features as a high-dimensional treatment within a causal framework. 
Based on the node embedding of Graph Neural Network (GNN), CECF seeks to learn a balanced representation of high-dimensional edge features by mitigating the potential influence of node features. 
Then, a cross-attention network captures the complex dependencies between node and edge features for final edge classification.
Extensive experiments demonstrate that CECF not only achieves superior performance but also serves as a flexible, plug-and-play enhancement for existing methods.
We also provide empirical analyses, offering insights into when and how this high-dimensional causal modeling framework works for the edge classification.
\end{abstract}

\section{Introduction}
\label{Introduction}
Edge classification, crucial for a variety of applications ranging from categorizing protein interactions to understanding social media connections \cite{pandey2019comprehensive,jha2022prediction}, remains surprisingly under-explored within the Graph Machine Learning community, particularly in comparison to the more widely studied  link prediction task \cite{wang2024topological,wang2024optimizing,zhao2022learning}.

Existing approaches to edge classification primarily address two key challenges.  First, they leverage Graph Neural Networks (GNNs) to learn node and edge embeddings that effectively capture graph structure and node/edge features \cite{kipf2016semi,kim2019edge}.  Many such methods incorporate techniques like attention mechanisms \cite{velivckovic2017graph} and advanced approximation operators \cite{tang2024chebnet}, proven successful in other domains. Second, they tackle the class imbalance problem inherent in many edge classification tasks, aiming to mitigate biases introduced by skewed edge distributions \cite{zhu2021graph,cheng2024edge}.  However, these approaches often overlook the crucial interplay between nodes and edges, relying on simpler operations like concatenation or aggregation of their embeddings.  This simplification potentially leads to information loss by neglecting the mutual influence between nodes and their connecting edges.

Consider, for example, a NetFlow-based IoT network \cite{sarhan2021netflow}. In this context, node represents features like user IP addresses, while edge could include features like flow duration, flow bytes, and packets. We want to classify each edge as belonging to one of several categories, such as injection, DDoS attack, benign, or scanning. It becomes apparent that while the final edge classification is influenced by both node and edge features, the edge features are also inherently affected by the connected nodes. Consequently, simple concatenation or aggregation techniques cannot fully capture or mitigate the potential relationships between these high-dimensional node features and edge features, leading to associated noise and a loss of valuable information. 

To investigate how to effectively leverage the interplay between nodes and edges for edge classification, we explore modeling this interplay within a Causal Edge Classification Framework (CECF), specifically designed by addressing three key challenges inherent in applying causal inference to this high-dimensional setting. For the challenge of \textbf{Causal Graph Construction} in edge classification, our CECF defines node features as covariates, edge features as a high-dimensional treatment, and the edge class as the target, explicitly modeling potential confounding. For the challenge of selecting a suitable \textbf{Causal Inference Framework} for high-dimensional treatments in edge classification where counterfactual generation is difficult, we adopt an adversarial approach \cite{kazemi2024adversarially} to learn a balanced representation of edge features by training a GNN for node embeddings alongside a model to estimate edge features, thereby mitigating confounding from nodes. Lastly, for \textbf{Extending Causal Frameworks to High-Dimensional Treatments}, unlike traditional models typically designed for one-dimensional treatments \cite{wei2024multi,kazemi2024adversarially}, CECF incorporates modeling techniques, such as employing network-based approximations rather than one-dimensional spline and formulating the high-dimensional regression problem, to effectively process and represent the entire high-dimensional edge feature vector. After addressing these challenges, these learned node and balanced edge embeddings are then fed into a cross-attention network to capture their complex dependencies for the final edge classification. To our best knowledge, CECF is the first to explore such a comprehensive extension for causal edge classification considering high-dimensional node-edge interplay.
 
We empirically validated our CECF on various datasets, observing substantial improvements in edge classification compared to existing methods in most cases, showing effective both as a standalone model and a plug-and-play enhancement within an acceptable computational overhead. To deeply understand when and how CECF works effectively, we conducted extensive additional experiments and case studies. Our analysis reveals that CECF particularly excels when strong causal relationships exist, specifically where node features directly influence edge features. In such scenarios, CECF not only establishes stronger representations for both nodes and edges that are more indicative of the final class, but also more effectively manages the combined contribution of these representations to the ultimate prediction. These insights aim to identify promising avenues for applying causal inference models to edge classification tasks involving high-dimensional treatments.

Our contributions are summarized as follows: (1) We highlight the critical edge classification task, emphasizing the impact of interplay between nodes and their connecting edges on edge classification outcomes. (2) We propose the Causal Edge Classification Framework (CECF), a novel approach that models the high-dimensional causal relationships between node features, edge features, and edge classes. It is the first for the edge classification task to explore modeling edge features as a high-dimensional treatment within a causal framework. (3) We demonstrate through extensive experiments that CECF outperforms existing methods in most cases and can serve as a plug-and-play module, while we also identify and analyze when and how does the node features causally influence edge features.

\section{Problem Definition}\label{sec:Problem Definition}
\textbf{Notation}\quad Let $G = (\mathcal{V}, \mathcal{E}, \mathbf{X}, \mathbf{S})$ be an undirected attributed graph. 
The $\mathcal{V} = \{v_i\}_{i = 1}^{n}$ represents the set of $n$ nodes, where $n = |\mathcal{V}|$. Based on this, we denote $\mathcal{N}_i$ as the set of neighbor nodes for node $v_i$.
The $\mathcal{E} \subseteq \mathcal{V} \times \mathcal{V}$ represents the set of $m$ observed edges, where $m = |\mathcal{E}|$. Based on this, we denote an edge between node $v_i$ and $v_j$ as $e_{ij}\in \mathcal{E}$, and they further compose the adjacency matrix $\mathbf{A} \in \{0, 1\}^{n \times n}$. $\mathbf{A}_{ij} = 1$ if an edge exists between node $v_i$ and $v_j$, and $\mathbf{A}_{ij} = 0$ otherwise.
The $\mathbf{X} \in \mathbb{R}^{n \times d_1}$ is the node feature matrix, where each node is represented by a $d_1$-dimensional feature vector. 
The $\mathbf{S} \in \mathbb{R}^{n \times n \times d_2}$ is the edge feature matrix, where each edge is represented by a $d_2$-dimensional feature vector. When $\mathbf{A}_{ij} = 1$, the tensor element $\mathbf{S}_{ij}$ contains meaningful information.

\noindent\textbf{Problem Statement}\quad In this work, we follow the commonly accepted problem definition of edge classification \cite{aggarwal2016edge,cheng2024edge}. Given a graph $G = (\mathcal{V}, \mathcal{E}, \mathbf{X}, \mathbf{S})$ and the corresponding edge class labels $\mathbf{Y} \in \{0, 1\}^{m \times C}$ ($C$ is the total number of edge classes), we need to learn an edge classifier based on the edge class labels in the training set $\mathbf{Y}^{\text{Tr}}$ to accurately classify the edge class labels in the validation and testing sets $\mathbf{Y}^{\text{Val/Test}}$. In particular, we aim to leverage the node and edge features, and the graph structure itself, to predict the class membership of each edge.

\section{Method}
This section presents our Causal Edge Classification Framework (CECF) with high-dimensional treatments. To effectively leverage the interplay between nodes and edges within a causal inference model and easy to integrate with current graph models, our framework needs to address three key challenges:
(1) \textbf{Causal Graph Construction:} How to construct a causal graph that captures the interplay between nodes and edges? 
(2) \textbf{Causal Inference Framework Selection:} Which causal inference framework is easy and best suited for exploring node and edge features with high-dimensions, especially when generating counterfactual samples becomes significantly more challenging due to the high dimensionality and uncertain relationships between features?
(3) \textbf{Extending Causal Frameworks to High-Dimensional Treatments:}  Even if we choose a causal inference framework, which has potential to handle the high-dimensional treatments, existing models are still designed for one dimensional treatments. How can we design specific details of the framework to efficiently accommodate high-dimensional treatments while maintaining computational feasibility?

\begin{figure*}[h]
  \centering
  \includegraphics[width=0.7\linewidth]{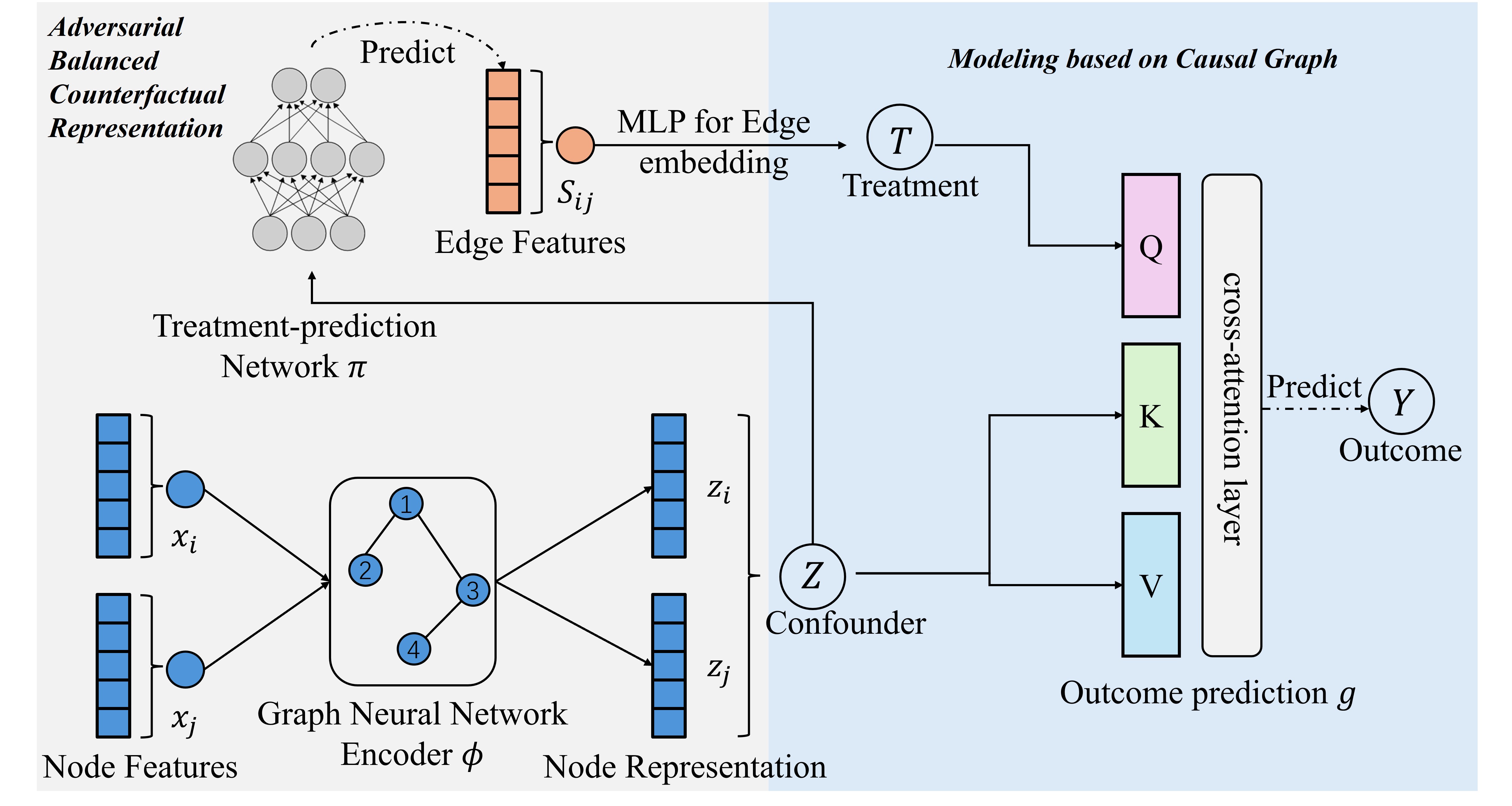}
  \caption{Overview of the Causal Edge Classification Framework (CECF). It leverages adversarial learning within a causal framework for edge classification, explicitly modeling the interplay between high-dimensional nodes and edges features.}
  \label{sec3}
\end{figure*}

To address these challenges, our CECF comprises three key components: Causal Graph Construction, Adversarial Balanced Counterfactual Representation on Graphs, and High-Dimensional Treatment Processing. The overall architecture is illustrated in Figure \ref{sec3}. The subsequent sections will detail each component.

\subsection{Causal Graph Construction}
To address the first challenge, the first step in our CECF is to construct a causal graph that explicitly represents the relationships between node features, edge features, and the target edge class. 

\begin{figure}[h]
  \centering
  \includegraphics[width=0.8\linewidth]{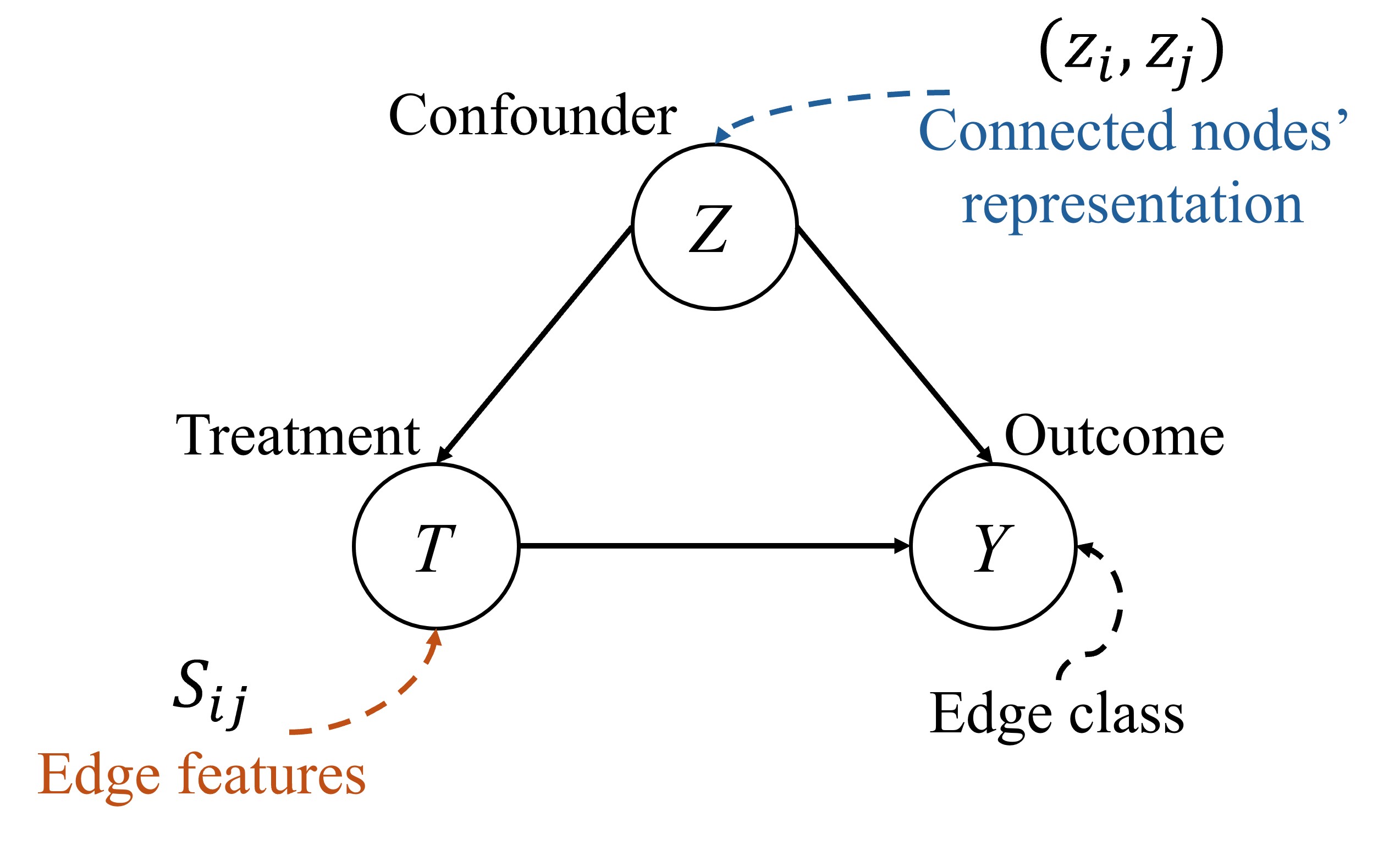}
  \caption{A Basic Causal Graph and Node-Edge Interplay Modeling.}
  \label{sec31}
\end{figure}

In a causal graph, it typically comprises 3 core components: the context (confounder) $Z$, the treatment $T$, and the outcome $Y$ \cite{morgan2015counterfactuals}, shown in Figure \ref{sec31}. In this graph, given the context, treatments, and their corresponding outcomes, we aim to identify the causal effects of both the context and the treatment on the outcome.

In our specific setting of edge classification with interplay between nodes and edges, the goal is to learn effective node representations for nodes $\mathcal{V}$ and to leverage edge features $\mathbf{S}$ for accurate edge classification $\mathbf{Y}$ in test data. Drawing upon the example discussed in the Section \ref{Introduction}, we recognize that node features often exist prior to the formation of edges between nodes. Consequently, it gives a hypothesis that node features frequently influence the characteristics of the edges connecting them.

Therefore, within our causal graph framework, the context $Z$ represents the node representations.  Because an edge is determined by its two connected nodes, $Z$ encompasses the representations $z_i$ and $z_j$ of two connected nodes $v_i$ and $v_j$, respectively.  The edge feature $\mathbf{S}_{ij}$ connecting these nodes serves as the treatment $T$, and the edge's class constitutes the outcome $Y$. Consequently, our causal graph captures the potential confounding effect of node features $Z$ on the relationship between edge features $T$ and edge class $Y$.

\subsection{Adversarial Balanced Counterfactual Representation on Graphs}
\label{sebsec:Adversarial}
Given the constructed causal graph and to address the second challenge, we aim to find a related causal model. The model needs to mitigate the confounding effects of node features on the relationship between edge features and the target edge class and further predict the edge class based on the node and edge features. To achieve this, prior approaches often relied on directly constructing counterfactual samples to augment the training data \cite{zhao2022learning,guo2025counterfactual}. This necessitates defining how treatments are perturbed or altered, which is readily achievable in one-dimensional, discrete, or even continuous settings. However, this becomes significantly more challenging for high-dimensional treatments, like our edge feature $\mathbf{S}_{ij}$. \footnote{Since each dimension of edge features is not in the same space, and the relationships between them are uncertain, dimensional reduction methods like clustering \cite{rokach2005clustering} or PCA \cite{abdi2010principal} are not suitable.} 
Therefore, we seek a causal learning framework that circumvents the direct construction of counterfactual samples.

Inspired by the work of \cite{kazemi2024adversarially}, which leverages adversarial learning to achieve counterfactual representation in a one-dimensional continuous treatment setting, we adapt this approach to our graph and extend it to handle high-dimensional treatments. Similar to their approach, we aim to control the counterfactual error caused by confounding effects by learning an encoder function $\phi$ and an outcome prediction function $g$ that jointly minimize \textbf{a distribution shift} and \textbf{outcome prediction error}. 
The minimizing of distribution shift is aimed at reducing the divergence between the joint distribution of $Z$ and $T$, denoted as $p(Z, T)$, and the product of their marginal distributions $p(Z)p(T)$. This effectively promotes independence between $Z$ and $T$, thereby mitigating the influence of confounding effects.

In our graph scenario, the encoder function $\phi$ refers to the node representation generation model $z_i = \phi(x_i)$, which naturally lends to using a Graph Neural Network (GNN). The prediction function $g$ corresponds to the model that uses the paired node representations $Z=(z_i, z_j)$ and edge features $T$ to perform outcome prediction. Based on prior studies \cite{zhang2022exploring,wei2024multi}, a cross-attention layer followed by a linear layer is likely to be effective for this task, which can model interactions between treatments $T$ and covariates $Z$. The cross-attention layer utilizes three matrices: a query $Q$, a key $K$, and a value $V$. In our case, $Q$ is learned from the treatment embedding using parameters $g_q$, while $K$ and $V$ are learned from the node representations using parameters $g_k$ and $g_v$, respectively.

Therefore, to \textbf{minimize distribution shift}, we minimize the KL divergence between $p(Z, T)$ and $p(Z)p(T)$. Following the derivation in \cite{kazemi2024adversarially}, this aim is to maximize the conditional entropy $H(T| (\phi(x_i), $ $ \phi(x_j))) = -\mathbb{E}_{p(T,Z)}[\log p(T|Z)]$. As $p(T|Z)$ is unknown, we also introduce a variational distribution $q_{\pi}(T|Z)$ to approximate it. We assume that the distribution $q_{\pi}(T|Z)$ is a Gaussian distribution with a fixed variance. Then, we can estimate the mean of $q_\pi(T|Z)$ using a neural network, which we refer to as the treatment-prediction network $\pi$. The minimization objection can be expressed as:

\begin{align}
  &\max_{\phi}H(T| (\phi(x_i), \phi(x_j))) = \nonumber\\
    &\max_{\phi}\inf_{\pi} \mathbb{E}_{p(S_{ij},  (\phi(x_i), \phi(x_j)))}[-\log q_{\pi}(T| (\phi(x_i), \phi(x_j)))].
    \label{eq1:begin}
\end{align}

Ultimately, from this, we construct the following loss function to encourage the counterfactual
representation:
\begin{equation}
        \max_{\phi} \min_{\pi} \mathcal{L}_{rep} = \frac{1}{|\mathcal{D}|}\sum_{(i, j) \in \mathcal{D}}  ( \mathbf{S}_{ij} - \pi(\phi(x_i), \phi(x_j))) ^2, 
    \label{eq3}
\end{equation}
where $\mathcal{D}$ represents the training set.

For \textbf{minimizing outcome prediction error}, we directly optimize the performance of the outcome prediction function $g$ in predicting $Y$ from the paired node representations $Z$ and the edge features $T$. This corresponds to the edge classification model optimization step, which can be formally expressed through the following cross-entropy loss function:
\begin{equation}
    \min_{\phi, g} \mathcal{L}_{pre} =  -\frac{1}{|\mathcal{D}|} \sum_{(i, j) \in \mathcal{D}} \sum_{c=1}^{C} Y_{ij}^c \log \left( g_c(\phi(x_i), \phi(x_j), \mathbf{S}_{ij}) \right).
\label{eq4}
\end{equation}

In this Equation, $C$ denotes the total number of edge classes, and $g_c$ represents the probability of prediction function $g$ predicting class $c$, addressing the multi-class classification challenge. Additionally, the parameters of the encoder model $\phi$ can be optimized to further improve the model's performance.

Combining these two objectives, we obtain our end-to-end loss function for the adversarial balanced counterfactual representation:
\begin{align}
   \min_{\phi, g} \max_{\pi} \mathcal{L} =  \mathcal{L}_{pre} - \gamma \mathcal{L}_{rep},
   \label{eq5}
\end{align}
where $\gamma$ is a hyperparameter. During training, on each data batch, we first update the treatment-prediction network $\pi$, and then update the encoder model $\phi$ and the prediction function $g$. We train our framework by alternating two steps in each batch: first, updating a network ($\pi$) to predict edge features of the node embeddings, and second, updating the main model ($\phi$ and $g$) to perform well in the classification task while making its node embeddings less predictive of edge features. \footnote{The pseudocode of CECF is provided in Appendix~\ref{appendix:pseudocode}.}

\subsection{High-Dimensional Treatment Processing}
\label{subsec:High-Dimensional}
While the aforementioned design and training objectives ensure that we can perform causal edge classification without directly constructing counterfactual samples, certain implementation details require further design when extending the framework from low-dimensional to high-dimensional treatments. Therefore, we introduce additional High-Dimensional Treatment Processing steps to guarantee the feasibility and effectiveness of the overall framework, which address the third challenge.

First, the treatment $T$ (or equivalently, the edge feature $\mathbf{S}_{ij}$) in the previous formulations is now a high-dimensional vector, unlike the one-dimensional treatments considered in prior work. This implies that the Equation (\ref{eq3}) needs to be rewritten. Consequently, we adjust the loss function to use the 2-norm to measure the difference between the estimated and observed edge features \footnote{We provide a theoretical guarantee based on the concentration of measure phenomenon \cite{vershynin2018high} in Appendix \ref{app:theory}.}:

\begin{equation}
        \max_{\phi} \min_{\pi}  \mathcal{L}_{rep} = \frac{1}{|\mathcal{D}|} \sum_{(i, j) \in \mathcal{D}}  ||\mathbf{S}_{ij} - \pi(\phi(x_i), \phi(x_j))||_2 ^2.
        \label{eq5:important}
\end{equation}

Second, in previous one-dimensional treatment settings, the treatment variable $\mathbf{S}_{ij}$ would typically be processed using spline functions to obtain an initial embedding before being fed into the prediction function $g(\phi(x_i), \phi(x_j), \mathbf{S}_{ij})$. This approach leverages piecewise function approximation to enhance the continuity and information content of the one-dimensional treatment. However, this technique becomes unsuitable when $\mathbf{S}_{ij}$ is high-dimensional. In particular, the uncertain relationships between the dimensions of our high-dimensional $\mathbf{S}_{ij}$ invalidate the assumptions underlying many high-dimensional interpolation algorithms, which often assume statistical homogeneity across dimensions. Therefore, we directly process $\mathbf{S}_{ij}$ through a neural network to generate the initial embedding. 

\begin{table*}[htbp]
\renewcommand{\arraystretch}{0.6}
  \centering
  \caption{Binary-class Edge Classification Performance on Various Datasets (Mean Values). The best performance is highlighted in bold. +CECF(TopoEdge) indicates that CECF is applied with the TopoEdge framework. For brevity, the complete results with standard deviations can be found in Appendix \ref{The Detail Results of The Main Results}.}
  \setlength{\tabcolsep}{0.8mm} 
  \small
  \begin{tabular}{ccccccccccccc}
    \toprule
    Dataset & \multicolumn{3}{c}{Bitcoin-alpha} & \multicolumn{3}{c}{HSPPI} & \multicolumn{3}{c}{Epinions} & \multicolumn{3}{c}{Reddit} \\
    \midrule
    Model & BACC  & Macro F1 & Time (S) & BACC  & Macro F1 & Time (S) & BACC  & Macro F1 & Time (S) & BACC  & Macro F1 & Time (S) \\
    \midrule
    TER+AER(Poisson) & 0.822 & 0.833 & 23.997 & 0.537 & 0.508 & 123.106 & 0.734 & 0.776 & 62.362 & 0.721 & 0.763 & 217.828 \\
    TER+AER(Geometric) & 0.872 & 0.874 & 23.879 & 0.517 & 0.474 & 124.851 & 0.724 & 0.771 & 61.267 & 0.728 & 0.769 & 216.739 \\
    \midrule
    GCN   & 0.798 & 0.836 & 9.483 & 0.500 & 0.432 & 198.538 & \textbf{0.756} & 0.757 & 133.076 & 0.744 & 0.766 & 277.184 \\
    +TopoEdge & 0.862 & 0.859 & 12.553 & 0.510 & 0.465 & 279.191 & 0.747 & 0.764 & 179.555 & 0.679 & 0.673 & 397.097 \\
    +CECF & 0.866 & 0.867 & 24.726 & 0.577 & 0.577 & 285.963 & 0.733 & 0.760 & 376.499 & \textbf{0.772} & \textbf{0.776} & 479.069 \\
    +CECF(Topoedge) & \textbf{0.872} & \textbf{0.871} & 26.801 & \textbf{0.626} & \textbf{0.627} & 365.140 & 0.748 & \textbf{0.765} & 398.645 & 0.708 & 0.681 & 491.184 \\
    \midrule
    GAT   & 0.826 & 0.846 & 12.076 & 0.500 & 0.433 & 201.142 & 0.726 & 0.751 & 158.308 & 0.539 & 0.536 & 218.781 \\
    +TopoEdge & 0.870 & 0.855 & 16.874 & 0.500 & 0.433 & 277.797 & 0.747 & 0.760 & 183.237 & 0.500 & 0.475 & 245.266 \\
    +CECF & 0.857 & 0.865 & 23.392 & 0.511 & 0.488 & 292.048 & 0.657 & 0.693 & 390.158 & \textbf{0.777} & \textbf{0.776} & 504.268 \\
    +CECF(Topoedge) & \textbf{0.872} & \textbf{0.870} & 23.262 & \textbf{0.543} & \textbf{0.533} & 372.402 & \textbf{0.750} & \textbf{0.763} & 413.414 & 0.559 & 0.544 & 589.884 \\
    \midrule
    ChebNet & 0.819 & 0.849 & 11.989 & 0.500 & 0.432 & 190.256 & 0.762 & 0.778 & 114.461 & 0.751 & 0.772 & 343.553 \\
    +TopoEdge & \textbf{0.882} & 0.871 & 21.697 & 0.500 & 0.432 & 272.613 & 0.760 & \textbf{0.784} & 171.193 & 0.682 & 0.678 & 394.174 \\
    +CECF & 0.850 & 0.866 & 23.433 & 0.640 & 0.640 & 273.911 & 0.758 & \textbf{0.784} & 340.686 & \textbf{0.795} & \textbf{0.784} & 367.216 \\
    +CECF(Topoedge) & 0.877 & \textbf{0.874} & 26.318 & \textbf{0.674} & \textbf{0.657} & 350.087 & \textbf{0.764} & 0.721 & 346.676 & 0.777 & 0.763 & 478.361 \\
    \bottomrule
  \end{tabular}
  \label{Binary}
\end{table*}%

\section{Experiments}\label{sec:Experiments}
In this section, we present a comprehensive evaluation of the proposed Causal Edge Classification Framework (CECF). We compare CECF against several baselines, demonstrating the importance of explicitly modeling the interplay between node and edge features. Furthermore, we provide more experiments to investigate when and how CECF works. \footnote{We also conducted the ablation studies of $\mathcal{L}_{rep}$ and the hyperparameter $\gamma$ of Equation (\ref{eq5}) in Appendix \ref{appendix:The Ablation of loss} and \ref{The Ablation of gamma}, a test of other statistical causal frameworks in Appendix \ref{app:Causal Frameworks}, and a in-depth case study and its analysis in Appendix \ref{app:case study}. }

\subsection{Experimental Setup}\label{sec:Experimental Setup}

\textbf{Datasets}\quad We evaluate the proposed CECF on six benchmark datasets, first introduced in~\cite{cheng2024edge}, each containing node features, edge features, and edge class labels. The datasets span various domains, including social trust networks (\textbf{Bitcoin-alpha} \cite{derr2017signed}, \textbf{Epinions} \cite{guha2004propagation}), a protein interaction network (\textbf{HSPPI} \cite{szklarczyk2023string}), a computer network (\textbf{NID} \cite{sarhan2021netflow}), a hyperlink network (\textbf{Reddit} \cite{kumar2018community}), and a citation network (\textbf{MAG} \cite{sinha2015overview}). For our experiments, Bitcoin-alpha, HSPPI, Epinions, and Reddit are framed as binary classification tasks, while Bitcoin-alpha (with different labels), NID, and MAG are used for multi-class classification.%
\footnote{Detailed description of datasets is provided in Appendix~\ref{app:datasets}.}

\textbf{Baselines}\quad To demonstrate the effectiveness, we compare CECF against several baseline methods:
(1) \textbf{GCN} \cite{kipf2016semi}, \textbf{GAT} \cite{velivckovic2017graph}, and \textbf{ChebNet} \cite{tang2024chebnet}:  These are fundamental Graph Neural Networks (GNNs) used to learn node embeddings. When using these models, we compute node representations by them, incorporate edge features, and then use an MLP to predict the final edge class.
(2) \textbf{TopoEdge} \cite{cheng2024edge}:  A recent state-of-the-art model leveraging topological information for edge classification. It modifies the loss function to incorporate topology-aware weighting and sampling on the GNN backbone.
(3) \textbf{TER+AER} \cite{wang2023efficient}: Another state-of-the-art but non-GNN-based model that utilizes graph representation learning techniques to address the edge classification task, including TER+AER (Poisson) and TER+AER (Geometric) 2 versions.

\textbf{Implementation Details}\quad Following the benchmark construction in \cite{cheng2024edge}, we use a 1:2:2 ratio for training, validation, and testing sets across all datasets. Our framework also utilizes GCN, GAT, and ChebNet as backbones for node representations. 
As CECF is not in conflict with TopoEdge, we also explore combining CECF with TopoEdge. It shows the plug-and-play ability of our model.
The optimal value of $\gamma$ for balancing the Adversarial Balanced Counterfactual Representation and learning rate for $\pi$ are selected based on the validation set. All experiments were repeated three times to calculate the variance, which is used to evaluate the stability of the results. \footnote{For more details, including dimensional settings, the integration of CECF with TopoEdge, please refer to the Appendix \ref{appendix:Implementation Details of Main Results}.}

\begin{table*}[htbp]
\renewcommand{\arraystretch}{0.8}
  \centering
  \caption{Multi-class Edge Classification Performance on Various Datasets (Mean Values). The best performance is highlighted in bold. +CECF(TopoEdge) indicates that CECF is applied with the TopoEdge framework. The complete results with standard deviations are shown in Appendix \ref{The Detail Results of The Main Results}.}
     \setlength{\tabcolsep}{0.8mm} 
     \small
     \begin{tabular}{cccccccccc}
    \toprule
    Dataset & \multicolumn{3}{c}{Bitcoin-alpha} & \multicolumn{3}{c}{NID} & \multicolumn{3}{c}{MAG} \\
    \midrule
    Model & BACC  & Macro F1 & Time(S) & BACC  & Macro F1 & Time (S) & BACC  & Macro F1 & Time (S) \\
    \midrule
    TER+AER(Poisson) & 0.364 & 0.392 & 23.913 & 0.324 & 0.329 & 62.122 & 0.643 & 0.704 & 44.780 \\
    TER+AER(Geometric) & 0.486 & 0.507 & 24.410 & 0.319 & 0.326 & 41.928 & 0.638 & 0.701 & 44.909 \\
    \midrule
    GCN   & 0.514 & 0.531 & 15.209 & 0.523 & 0.510 & 148.578 & 0.797 & 0.834 & 95.890 \\
    +TopoEdge & 0.572 & 0.558 & 12.369 & 0.627 & 0.491 & 200.487 & \textbf{0.808} & \textbf{0.840} & 129.986 \\
    +CECF & 0.537 & 0.534 & 24.936 & 0.523 & 0.507 & 199.373 & 0.694 & 0.733 & 158.244 \\
    +CECF(Topoedge) & \textbf{0.577} & \textbf{0.564} & 29.585 & \textbf{0.733} & \textbf{0.555} & 228.624 & 0.767 & 0.813 & 189.877 \\
    \midrule
    GAT   & 0.524 & 0.544 & 18.896 & 0.495 & 0.490 & 152.463 & 0.795 & 0.837 & 120.894 \\
    +TopoEdge & 0.577 & \textbf{0.565} & 16.713 & 0.558 & 0.484 & 206.151 & \textbf{0.812} & \textbf{0.840} & 146.772 \\
    +CECF & 0.538 & 0.552 & 32.554 & 0.521 & 0.507 & 215.282 & 0.732 & 0.756 & 201.003 \\
    +CECF(Topoedge) & \textbf{0.579} & 0.548 & 39.390 & \textbf{0.693} & \textbf{0.544} & 242.679 & 0.790 & 0.810 & 237.027 \\
    \midrule
    ChebNet & 0.528 & 0.557 & 15.965 & 0.527 & 0.514 & 167.421 & 0.801 & \textbf{0.842} & 89.985 \\
    +TopoEdge & \textbf{0.588} & 0.578 & 19.352 & 0.651 & 0.522 & 194.287 & \textbf{0.802} & 0.833 & 119.371 \\
    +CECF & 0.526 & 0.559 & 26.139 & 0.527 & 0.514 & 196.057 & 0.722 & 0.732 & 101.261 \\
    +CECF(Topoedge) & 0.585 & \textbf{0.583} & 29.972 & \textbf{0.683} & \textbf{0.561} & 219.061 & 0.776 & 0.801 & 170.096 \\
    \bottomrule
    \end{tabular}
  \label{Multi}%
\end{table*}%

\textbf{Evaluation Metrics}\quad  We use balanced accuracy (BACC) and Macro-F1 (Macro F1) as evaluation metrics, which are well-suited for assessing performance on imbalanced datasets, as is the case in this benchmark. Additionally, we measure training time to evaluate whether our CECF framework introduces significant overhead when modifying existing baseline models.

\subsection{Main Results}\label{sec:Main Results}
The results in Tables \ref{Binary} and \ref{Multi} demonstrate the significant impact of our Causal Edge Classification Framework (CECF). Our findings highlight two key aspects: CECF's effectiveness in enhancing basic GNN models and its ability to augment current state-of-the-art methods by addressing the causal problem of edge classification.

Firstly, when applied to basic GNN models, \textbf{CECF achieves notable standalone gains, typically ranging from 2-5\% in BACC and/or Macro-F1 over these backbones on most datasets.}\footnote{With the Wilcoxon signed-rank test, it is a statistically significant difference ($p < 0.05$) between our model and other baseline.} For instance, in binary classification on the Reddit dataset, applying CECF to the GCN backbone substantially improved performance from a 0.744 BACC score and 0.766 Macro F1 score to 0.772 BACC and 0.776 Macro F1, respectively. Similarly, for the NID multi-class dataset, integrating CECF with the GAT backbone increased the BACC score from 0.495 to 0.521 and the Macro F1 score from 0.490 to 0.507. This consistent positive uplift across different GNN backbones and settings validates our hypothesis: explicitly modeling the causal node-edge interplay captures significant, previously unaddressed information, leading to more accurate edge classification.

Secondly, it is crucial to recognize that edge classification is a multifaceted challenge, \textbf{our framework can be a plug-and-play strategy to other models.} While existing state-of-the-art methods like TopoEdge effectively address specific issues such as topological class imbalance, our work is the first to investigate and model the high-dimensional causal relationships between node and edge, which further improves the performance of TopoEdge. This distinct contribution is evident when CECF augments state-of-the-art methods. Combining CECF with TopoEdge (\texttt{+CECF(TopoEdge)}) consistently outperforms TopoEdge alone, often by 1-10\%, establishing new state-of-the-art results on most datasets. For example, on the NID multi-class benchmark, \texttt{+CECF(TopoEdge)} (GCN) lifted BACC from 0.627 to 0.733. The enhanced results affirm that our approach augments existing methods, yielding more refined and accurate edge classifications by tackling the previously overlooked causal aspect.

Finally, regarding practical utility, \textbf{integrating CECF does not introduce prohibitive computational overhead.} CECF's (alone) training time is often comparable to TopoEdge (e.g., $\sim$280s on HSPPI, $\sim$200s on NID), with increases on other datasets (e.g., Reddit, Epinions) typically under 2x. While we find that non-GNN based methods, TER+AER, inherently possess a time advantage, their classification performance in our experiments does not match that of GNN-based approaches, including ours.\footnote{In Appendix \ref{app:Causal Frameworks}, we show CECF's performance and time both are  better than other causal frameworks.}

\subsection{Further Experiments}
\label{sec:further_analysis}
This section delves into a detailed analysis of CECF, aiming to elucidate the factors contributing to its success across most datasets while addressing its conditions, particularly the performance decrease on the MAG dataset. First, we assess the degree of interplay between node and edge features across different datasets. Second, we quantify the importance of node representations and edge features on the final prediction before and after modeling this interplay with CECF. The first evaluation aims to demonstrate the prevalence and significance of node-edge interplay in the datasets and to verify that our method can effectively mitigate such dependencies. The second evaluation provides insights into how CECF's modeling of node-edge interplay specifically influences the edge classification model's reliance on node and edge features. 

\noindent\textbf{Settings}\quad
Specifically, \textit{for the evaluation of interplay between node and edge features}, we employ Canonical Correlation Analysis (CCA) to quantify the correlation between node representations and edge features for models with and without CECF. The CCA is a multivariate statistical method that identifies relationships between two sets of variables by finding pairs of canonical variables that maximize their correlation \cite{hardoon2004canonical}.  \textit{For the evaluation of features' importance}, we calculate the average Shapley values across all dimensions for both node representations and edge features, reflecting the contribution to the final prediction \cite{sundararajan2020many}. \footnote{For implementation settings, refer to Appendix \ref{appendix:Implementation Details of Further Experiences}. Notably, since no prior works have conducted such analyses of high-dimensional causal inference, we are the first to extend such analytical tools to high-dimensional for a deeper understanding of CECF. } 

The results in Table~\ref{tab:cca_shapley_main} first validate the central hypothesis of this work: \textbf{in most edge classification datasets, node representations and edge features exhibit strong correlations and may potentially form causal relationships.} Across the datasets, baseline models (e.g., GCN, ChebNet) consistently show high CCA values, indicating significant interplay between node and edge features. For instance, in datasets such as Bitcoin-alpha, Epinions, and NID, the CCA values of GCN are particularly high, reaching 0.570, 0.803, and 0.911, respectively. This strong correlation underscores the necessity of explicitly modeling this interplay, as it is a prevalent characteristic of edge classification tasks. Although this correlation may not directly imply causation \cite{pearl2009causal}, our analysis in Section \ref{Introduction} suggests that causal relationships are highly plausible as a modeling framework for such scenarios. 

\begin{table}[htbp]
\renewcommand{\arraystretch}{0.6}
  \centering
  
  \caption{Comparison of CCA and Shapley Values for baseline models and our final CECF model. The complete results are shown in Appendix \ref{The Detail Results of The Further Experiences}.} 
  \label{tab:cca_shapley_main}
  \small
  \setlength{\tabcolsep}{0.8mm} 
  \begin{tabular}{cc|cc|cc}
    \toprule
          & Model & GCN   & +CECF & ChebNet & +CECF \\
    \midrule
    Dataset & Metrics & \multicolumn{4}{c}{Binary Classification} \\ 
    \midrule
    \multirow{3}[2]{*}{Bitcoin-alpha} & CCA          & 0.570  & 0.604  & 0.510  & 0.483  \\
          & Node Shapley & 0.0103 & 0.0480 & 0.0206 & 0.0733 \\
          & Edge Shapley & 0.0197 & 0.0254 & 0.0222 & 0.0187 \\
    \midrule
    \multirow{3}[2]{*}{Epinions}      & CCA          & 0.803  & 0.814  & 0.860  & 0.838  \\
          & Node Shapley & 0.0013 & 0.0030 & 0.0021 & 0.0009 \\
          & Edge Shapley & 0.0060 & 0.0042 & 0.0000 & 0.0000 \\
    \midrule
    Dataset & Metrics & \multicolumn{4}{c}{Multi-class Classification} \\ 
    \midrule
    \multirow{3}[2]{*}{NID}           & CCA          & 0.911  & 0.963  & 0.901  & 0.941  \\
          & Node Shapley & 0.0576 & 0.0925 & 0.0611 & 0.0404 \\
          & Edge Shapley & 1.0634 & 0.0265 & 0.8354 & 0.0301 \\
    \midrule
    \multirow{3}[2]{*}{MAG}           & CCA          & 0.156  & 0.324  & 0.078  & 0.165  \\
          & Node Shapley & 0.0006 & 0.0676 & 0.0005 & 0.0789 \\
          & Edge Shapley & 0.0612 & 0.0989 & 0.0550 & 0.1116 \\
    \bottomrule
\end{tabular}%
\end{table}

\textbf{CECF is effective in datasets with strong node-edge interplay, but its benefit diminishes in scenarios with weak interplay.} This is evident in datasets with high initial Canonical Correlation Analysis (CCA) scores like Bitcoin-alpha ($>0.5$ in GCN and ChebNet) and NID ($>0.8$), where CECF's causal modeling significantly enhances performance. Conversely, in datasets with weak interplay like MAG (CCA $<0.2$), CECF offers limited benefit. This highlights the specificity of our approach: CECF excels precisely because it tackles this confounding interplay, but consequently, it offers limited benefits when such confounding is weak or absent. \textit{Therefore, we also suggest a preliminary assessment for practitioners: a high initial CCA score ($>0.5$) on a simple GNN indicates strong confounding interplay, making the dataset a prime candidate for our CECF approach.}
More importantly, on datasets like NID, CECF corrects imbalanced feature reliance. For instance, with a GCN backbone, it curtails an over-reliance on edge features (Shapley drops from 1.0634 to 0.0265) while boosting node features, showing that \textbf{CECF learns a more balanced and effective representation of node and edge importance}.

\section{Related Work}\label{sec:Related Work}

\textbf{Edge Classification}\quad While related to tasks such as link prediction in Knowledge Graphs (KGs) \cite{shu2024knowledge,yang2024dual} or Signed Networks (SNs) \cite{hsieh2012low} which typically deals with binary (positive/negative) edge labels, edge classification aims to categorize existing edges into predefined classes based on the edges' features and the features of their incident nodes. It initially addressed by using structural similarity \cite{aggarwal2016edge}, has seen significant development in edge embedding techniques. These techniques can be categorized into shallow embedding and GNN-based approaches. The shallow embedding approaches, such as AttrE2vec~\cite{bielak2022attre2vec}, Edge2vec~\cite{wang2020edge2vec}, and TER+AER~\cite{wang2023efficient}, learn edge representations through techniques like random walks or autoencoders, primarily leveraging graph structure and features. The GNN-based methods, conversely, employ GNNs to generate node embeddings, from which edge embeddings are derived using operations like averaging or concatenation \cite{kipf2016semi,velivckovic2017graph,tang2024chebnet}, or to directly generate edge embeddings \cite{kim2019edge,gong2019exploiting}. Recently, some works have started to address the issue of class imbalance in edge classification. For instance, TOBA~\cite{liu2023topological} first approaches the problem from a topology-centric perspective. Moreover, TopoEdge \cite{cheng2024edge} further explores how topology-aware weighting and sampling strategies improve performance on common edge classification datasets and methods. However, these methods do not explicitly model the interplay between nodes and edges, which overlooks this crucial prior assumption.

\textbf{Causal Inference in Graphs}\quad Causal inference, vital for reasoning about data relationships beyond correlations \cite{kaddour2022causal,prado2024survey}, is increasingly applied to various graph tasks to enhance robustness, fairness, and interpretability, with applications in causal fairness, explanations, link/relation prediction, and recommendation \cite{guo2025counterfactual}. For instance, Causal Fairness mitigates attribute-based bias using counterfactual augmentations and minimizing representation discrepancies \cite{dai2022learning,zhu2023learning}. Causal Explanations improve interpretability by identifying minimal input changes altering model predictions \cite{ma2022clear,huang2023global}.
Causal Link and Relation Prediction endeavors to uncover causal relation of links or predict relations more robustly, for example, by generating counterfactual links/augmentations \cite{zhao2022learning}, identifying causal subgraphs for inductive relation prediction \cite{yu2024generalizable}, or using counterfactual learning for higher-order relation prediction \cite{guo2025counterfactual}. Causal Recommendation addresses biases using counterfactual reasoning or by modeling causal user/item feature-outcome relationships \cite{gao2024causal,xiao2022representation}. 
The successful application of causal modeling across these diverse tasks underscores the potential prevalence of underlying causal effects. However, their causal frameworks like Invariant Graph Learning \cite{sui2024invariant} or CausalGNN \cite{wang2022causalgnn} focus on node-level problems or rely on discrete environments. Our work is the first to apply causality to edge classification without such environmental constraints, making it practical for a broader class of graph problems.

\section{Conclusion}

We present the Causal Edge Classification Framework (CECF), a novel approach that explicitly models the interplay between high-dimensional node and edge features within a causal inference framework. Unlike existing methods overlooking the relationships between node and edge features, CECF treats edge features as high-dimensional treatments and employs adversarial learning to mitigate confounding effects, thereby addressing a critical limitation in edge classification with causal inference. 
Our experiments demonstrate that CECF serves as an effective plug-and-play framework and improves performance across a variety of datasets. 

Moreover, our further exploration highlights when and how CECF works. In datasets with weak node-edge interplay, like MAG, CECF's performance degrades as its assumed causal relationships may not align with the data structure, underscoring the need to evaluate dataset characteristics before applying causal methods. Furthermore, our case study shows that our method not only enhances the individual embeddings of nodes and edges but also improves the combined representation for more effective predictions.

\bibliographystyle{named}
\bibliography{ijcai26}

@article{pandey2019comprehensive,
  title={A comprehensive survey of edge prediction in social networks: Techniques, parameters and challenges},
  author={Pandey, Babita and Bhanodia, Praveen Kumar and Khamparia, Aditya and Pandey, Devendra Kumar},
  journal={Expert Systems with Applications},
  volume={124},
  pages={164--181},
  year={2019},
  publisher={Elsevier}
}

@article{jha2022prediction,
  title={Prediction of protein--protein interaction using graph neural networks},
  author={Jha, Kanchan and Saha, Sriparna and Singh, Hiteshi},
  journal={Scientific Reports},
  volume={12},
  number={1},
  pages={8360},
  year={2022},
  publisher={Nature Publishing Group UK London}
}

@inproceedings{wang2024topological,
  title={A topological perspective on demystifying gnn-based link prediction performance},
  author={Wang, Yu and Zhao, Tong and Zhao, Yuying and Liu, Yunchao and Cheng, Xueqi and Shah, Neil and Derr, Tyler},
  booktitle={ICLR},
  year={2024}
}

@inproceedings{wang2024optimizing,
  title={Optimizing Long-tailed Link Prediction in Graph Neural Networks through Structure Representation Enhancement},
  author={Wang, Yakun and Wang, Daixin and Liu, Hongrui and Hu, Binbin and Yan, Yingcui and Zhang, Qiyang and Zhang, Zhiqiang},
  booktitle={Proceedings of the 30th ACM SIGKDD Conference on Knowledge Discovery and Data Mining},
  pages={3222--3232},
  year={2024}
}

@inproceedings{zhao2022learning,
  title={Learning from counterfactual links for link prediction},
  author={Zhao, Tong and Liu, Gang and Wang, Daheng and Yu, Wenhao and Jiang, Meng},
  booktitle={International Conference on Machine Learning},
  pages={26911--26926},
  year={2022},
  organization={PMLR}
}

@article{kipf2016semi,
  title={Semi-supervised classification with graph convolutional networks},
  author={Kipf, Thomas N and Welling, Max},
  journal={arXiv preprint arXiv:1609.02907},
  year={2016}
}

@article{tang2024chebnet,
  title={ChebNet: efficient and stable constructions of deep neural networks with rectified power units via Chebyshev approximation},
  author={Tang, Shanshan and Li, Bo and Yu, Haijun},
  journal={Communications in Mathematics and Statistics},
  pages={1--27},
  year={2024},
  publisher={Springer}
}

@article{velivckovic2017graph,
  title={Graph attention networks},
  author={Veli{\v{c}}kovi{\'c}, Petar and Cucurull, Guillem and Casanova, Arantxa and Romero, Adriana and Lio, Pietro and Bengio, Yoshua},
  journal={arXiv preprint arXiv:1710.10903},
  year={2017}
}

@inproceedings{kim2019edge,
  title={Edge-labeling graph neural network for few-shot learning},
  author={Kim, Jongmin and Kim, Taesup and Kim, Sungwoong and Yoo, Chang D},
  booktitle={Proceedings of the IEEE/CVF conference on computer vision and pattern recognition},
  pages={11--20},
  year={2019}
}

@inproceedings{gong2019exploiting,
  title={Exploiting edge features for graph neural networks},
  author={Gong, Liyu and Cheng, Qiang},
  booktitle={Proceedings of the IEEE/CVF conference on computer vision and pattern recognition},
  pages={9211--9219},
  year={2019}
}

@inproceedings{cheng2024edge,
  author    = {Cheng, Xueqi and Wang, Yu and Liu, Yunchao and Zhao, Yuying and Aggarwal, Charu C and Derr, Tyler},
  title     = {Edge Classification on Graphs: New Directions in Topological Imbalance},
  booktitle = {WSDM},
  year      = {2025}
}

@inproceedings{zhu2021graph,
  title={Graph neural networks with heterophily},
  author={Zhu, Jiong and Rossi, Ryan A and Rao, Anup and Mai, Tung and Lipka, Nedim and Ahmed, Nesreen K and Koutra, Danai},
  booktitle={Proceedings of the AAAI conference on artificial intelligence},
  volume={35},
  number={12},
  pages={11168--11176},
  year={2021}
}

@inproceedings{sarhan2021netflow,
  title={Netflow datasets for machine learning-based network intrusion detection systems},
  author={Sarhan, Mohanad and Layeghy, Siamak and Moustafa, Nour and Portmann, Marius},
  booktitle={Big Data Technologies and Applications: 10th EAI International Conference, BDTA 2020, and 13th EAI International Conference on Wireless Internet, WiCON 2020, Virtual Event, December 11, 2020, Proceedings 10},
  pages={117--135},
  year={2021},
  organization={Springer}
}

@article{wei2024multi,
  title={Multi-Treatment Multi-Task Uplift Modeling for Enhancing User Growth},
  author={Wei, Yuxiang and Qiu, Zhaoxin and Li, Yingjie and Sun, Yuke and Li, Xiaoling},
  journal={arXiv preprint arXiv:2408.12803},
  year={2024}
}

@inproceedings{kazemi2024adversarially,
  title={Adversarially balanced representation for continuous treatment effect estimation},
  author={Kazemi, Amirreza and Ester, Martin},
  booktitle={Proceedings of the AAAI Conference on Artificial Intelligence},
  volume={38},
  number={12},
  pages={13085--13093},
  year={2024}
}

@inproceedings{aggarwal2016edge,
  title={Edge classification in networks},
  author={Aggarwal, Charu and He, Gewen and Zhao, Peixiang},
  booktitle={2016 IEEE 32nd international conference on data engineering (ICDE)},
  pages={1038--1049},
  year={2016},
  organization={IEEE}
}

@article{bielak2022attre2vec,
  title={Attre2vec: Unsupervised attributed edge representation learning},
  author={Bielak, Piotr and Kajdanowicz, Tomasz and Chawla, Nitesh V},
  journal={Information Sciences},
  volume={592},
  pages={82--96},
  year={2022},
  publisher={Elsevier}
}

@article{wang2020edge2vec,
  title={Edge2vec: Edge-based social network embedding},
  author={Wang, Changping and Wang, Chaokun and Wang, Zheng and Ye, Xiaojun and Yu, Philip S},
  journal={ACM Transactions on Knowledge Discovery from Data (TKDD)},
  volume={14},
  number={4},
  pages={1--24},
  year={2020},
  publisher={ACM New York, NY, USA}
}

@inproceedings{wang2023efficient,
  title={Efficient and effective edge-wise graph representation learning},
  author={Wang, Hewen and Yang, Renchi and Huang, Keke and Xiao, Xiaokui},
  booktitle={Proceedings of the 29th ACM SIGKDD Conference on Knowledge Discovery and Data Mining},
  pages={2326--2336},
  year={2023}
}

@article{liu2023topological,
  title={Topological augmentation for class-imbalanced node classification},
  author={Liu, Zhining and Zeng, Zhichen and Qiu, Ruizhong and Yoo, Hyunsik and Zhou, David and Xu, Zhe and Zhu, Yada and Weldemariam, Kommy and He, Jingrui and Tong, Hanghang},
  journal={arXiv preprint arXiv:2308.14181},
  year={2023}
}

@article{guo2025counterfactual,
  title={Counterfactual learning on graphs: A survey},
  author={Guo, Zhimeng and Wu, Zongyu and Xiao, Teng and Aggarwal, Charu and Liu, Hui and Wang, Suhang},
  journal={Machine Intelligence Research},
  volume={22},
  number={1},
  pages={17--59},
  year={2025},
  publisher={Springer}
}

@article{dai2022learning,
  title={Learning fair graph neural networks with limited and private sensitive attribute information},
  author={Dai, Enyan and Wang, Suhang},
  journal={IEEE Transactions on Knowledge and Data Engineering},
  volume={35},
  number={7},
  pages={7103--7117},
  year={2022},
  publisher={IEEE}
}

@article{zhu2023learning,
  title={Learning fair models without sensitive attributes: A generative approach},
  author={Zhu, Huaisheng and Dai, Enyan and Liu, Hui and Wang, Suhang},
  journal={Neurocomputing},
  volume={561},
  pages={126841},
  year={2023},
  publisher={Elsevier}
}

@inproceedings{huang2023global,
  title={Global counterfactual explainer for graph neural networks},
  author={Huang, Zexi and Kosan, Mert and Medya, Sourav and Ranu, Sayan and Singh, Ambuj},
  booktitle={Proceedings of the Sixteenth ACM International Conference on Web Search and Data Mining},
  pages={141--149},
  year={2023}
}

@article{ma2022clear,
  title={Clear: Generative counterfactual explanations on graphs},
  author={Ma, Jing and Guo, Ruocheng and Mishra, Saumitra and Zhang, Aidong and Li, Jundong},
  journal={Advances in neural information processing systems},
  volume={35},
  pages={25895--25907},
  year={2022}
}

@article{gao2024causal,
  title={Causal inference in recommender systems: A survey and future directions},
  author={Gao, Chen and Zheng, Yu and Wang, Wenjie and Feng, Fuli and He, Xiangnan and Li, Yong},
  journal={ACM Transactions on Information Systems},
  volume={42},
  number={4},
  pages={1--32},
  year={2024},
  publisher={ACM New York, NY}
}

@inproceedings{xiao2022representation,
  title={Representation matters when learning from biased feedback in recommendation},
  author={Xiao, Teng and Chen, Zhengyu and Wang, Suhang},
  booktitle={Proceedings of the 31st ACM International Conference on Information \& Knowledge Management},
  pages={2220--2229},
  year={2022}
}

@article{kaddour2022causal,
  title={Causal machine learning: A survey and open problems},
  author={Kaddour, Jean and Lynch, Aengus and Liu, Qi and Kusner, Matt J and Silva, Ricardo},
  journal={arXiv preprint arXiv:2206.15475},
  year={2022}
}

@article{prado2024survey,
  title={A survey on graph counterfactual explanations: definitions, methods, evaluation, and research challenges},
  author={Prado-Romero, Mario Alfonso and Prenkaj, Bardh and Stilo, Giovanni and Giannotti, Fosca},
  journal={ACM Computing Surveys},
  volume={56},
  number={7},
  pages={1--37},
  year={2024},
  publisher={ACM New York, NY}
}

@book{morgan2015counterfactuals,
  title={Counterfactuals and causal inference},
  author={Morgan, SL},
  year={2015},
  publisher={Cambridge University Press}
}

@article{rokach2005clustering,
  title={Clustering methods},
  author={Rokach, Lior and Maimon, Oded},
  journal={Data mining and knowledge discovery handbook},
  pages={321--352},
  year={2005},
  publisher={Springer}
}

@article{abdi2010principal,
  title={Principal component analysis},
  author={Abdi, Herv{\'e} and Williams, Lynne J},
  journal={Wiley interdisciplinary reviews: computational statistics},
  volume={2},
  number={4},
  pages={433--459},
  year={2010},
  publisher={Wiley Online Library}
}

@article{zhang2022exploring,
  title={Exploring transformer backbones for heterogeneous treatment effect estimation},
  author={Zhang, Yi-Fan and Zhang, Hanlin and Lipton, Zachary C and Li, Li Erran and Xing, Eric P},
  journal={arXiv preprint arXiv:2202.01336},
  year={2022}
}

@article{derr2017signed,
  title={Signed node relevance measurements},
  author={Derr, Tyler and Wang, Chenxing and Wang, Suhang and Tang, Jiliang},
  journal={arXiv preprint arXiv:1710.07236},
  year={2017}
}

@article{szklarczyk2023string,
  title={The STRING database in 2023: protein--protein association networks and functional enrichment analyses for any sequenced genome of interest},
  author={Szklarczyk, Damian and Kirsch, Rebecca and Koutrouli, Mikaela and Nastou, Katerina and Mehryary, Farrokh and Hachilif, Radja and Gable, Annika L and Fang, Tao and Doncheva, Nadezhda T and Pyysalo, Sampo and others},
  journal={Nucleic acids research},
  volume={51},
  number={D1},
  pages={D638--D646},
  year={2023},
  publisher={Oxford University Press}
}

@inproceedings{guha2004propagation,
  title={Propagation of trust and distrust},
  author={Guha, Ramanthan and Kumar, Ravi and Raghavan, Prabhakar and Tomkins, Andrew},
  booktitle={Proceedings of the 13th international conference on World Wide Web},
  pages={403--412},
  year={2004}
}

@inproceedings{kumar2018community,
  title={Community interaction and conflict on the web},
  author={Kumar, Srijan and Hamilton, William L and Leskovec, Jure and Jurafsky, Dan},
  booktitle={Proceedings of the 2018 world wide web conference},
  pages={933--943},
  year={2018}
}

@inproceedings{sinha2015overview,
  title={An overview of microsoft academic service (mas) and applications},
  author={Sinha, Arnab and Shen, Zhihong and Song, Yang and Ma, Hao and Eide, Darrin and Hsu, Bo-June and Wang, Kuansan},
  booktitle={Proceedings of the 24th international conference on world wide web},
  pages={243--246},
  year={2015}
}

@article{hardoon2004canonical,
  title={Canonical correlation analysis: An overview with application to learning methods},
  author={Hardoon, David R and Szedmak, Sandor and Shawe-Taylor, John},
  journal={Neural computation},
  volume={16},
  number={12},
  pages={2639--2664},
  year={2004},
  publisher={MIT Press}
}

@inproceedings{sundararajan2020many,
  title={The many Shapley values for model explanation},
  author={Sundararajan, Mukund and Najmi, Amir},
  booktitle={International conference on machine learning},
  pages={9269--9278},
  year={2020},
  organization={PMLR}
}

@article{pearl2009causal,
  title={Causal inference in statistics: An overview},
  author={Pearl, Judea},
  year={2009}
}

@article{shu2024knowledge,
  title={Knowledge graph large language model (KG-LLM) for link prediction},
  author={Shu, Dong and Chen, Tianle and Jin, Mingyu and Zhang, Chong and Du, Mengnan and Zhang, Yongfeng},
  journal={arXiv preprint arXiv:2403.07311},
  year={2024}
}

@inproceedings{hsieh2012low,
  title={Low rank modeling of signed networks},
  author={Hsieh, Cho-Jui and Chiang, Kai-Yang and Dhillon, Inderjit S},
  booktitle={Proceedings of the 18th ACM SIGKDD international conference on Knowledge discovery and data mining},
  pages={507--515},
  year={2012}
}

@article{yang2024dual,
  title={Dual De-confounded Causal Intervention method for knowledge graph error detection},
  author={Yang, Yunxiao and Chen, Jianting and Gao, Xiaoying and Xiang, Yang},
  journal={Knowledge-Based Systems},
  volume={305},
  pages={112644},
  year={2024},
  publisher={Elsevier}
}

@article{yu2024generalizable,
  title={Generalizable inductive relation prediction with causal subgraph},
  author={Yu, Han and Liu, Ziniu and Tu, Hongkui and Chen, Kai and Li, Aiping},
  journal={World Wide Web},
  volume={27},
  number={3},
  pages={24},
  year={2024},
  publisher={Springer}
}

@inproceedings{sui2024invariant,
  title={Invariant graph learning for causal effect estimation},
  author={Sui, Yongduo and Tang, Caizhi and Chu, Zhixuan and Fang, Junfeng and Gao, Yuan and Cui, Qing and Li, Longfei and Zhou, Jun and Wang, Xiang},
  booktitle={Proceedings of the ACM Web Conference 2024},
  pages={2552--2562},
  year={2024}
}

@inproceedings{wang2022causalgnn,
  title={Causalgnn: Causal-based graph neural networks for spatio-temporal epidemic forecasting},
  author={Wang, Lijing and Adiga, Aniruddha and Chen, Jiangzhuo and Sadilek, Adam and Venkatramanan, Srinivasan and Marathe, Madhav},
  booktitle={Proceedings of the AAAI conference on artificial intelligence},
  volume={36},
  number={11},
  pages={12191--12199},
  year={2022}
}

@book{vershynin2018high,
  title={High-dimensional probability: An introduction with applications in data science},
  author={Vershynin, Roman},
  volume={47},
  year={2018},
  publisher={Cambridge university press}
}

\clearpage
\appendix
\section{Implementation Details}
\label{appendix:Implementation Details}
In this section, we provide comprehensive details on the implementation of our framework and experiments. Specifically, we provide the pseudocode of CECF, the parameter settings and training procedures for the main results, and additional implementation details for the further experiments conducted in this work. All the experiments were conducted on a Linux server equipped with 2 Intel(R) Xeon(R) Gold 6348 CPU @ 2.60GHz, 1T of RAM, and 8 NVIDIA A100 cards(80GB of RAM each). Our methods were implemented using Python 3.10.16 alongside PyTorch 2.5.1+cu121. The operating system was Ubuntu20.04 with CUDA version 12.2.

\subsection{Pseudocode of CECF}\label{appendix:pseudocode}
To provide a comprehensive understanding of the framework's implementation, we now present the pseudocode for our CECF, detailing both the training and classification phases. Note that the processing of edge features $\mathbf{S}_{ij}$ using a neural network is implicitly integrated into the prediction function $g$, which is jointly updated during training.

\begin{algorithm}[H]
\renewcommand{\arraystretch}{0.6}
\caption{Causal Edge Classification Framework (CECF)}
\label{alg:cecf}
\begin{algorithmic}[1]
\Require Graph $G = (\mathcal{V}, \mathcal{E}, \mathbf{X}, \mathbf{S})$, edge labels $\mathbf{Y}$, hyperparameter $\gamma$, number of epochs $E$, each batch $B$, learning rate $\eta_{\pi}$ and $\eta$.
\State \textbf{Step 1: Causal Graph Construction}
\State Initialize Graph Neural Network encoder $\phi$.
\State Construct causal graph with node features $Z = (\phi(x_i), \phi(x_j))$ as context, edge features $T = \mathbf{S}_{ij}$ as treatment, and edge class $Y$ as outcome.
\State \textbf{Step 2: Adversarial Balanced Counterfactual Representation}
\State Initialize treatment-prediction network $\pi$ and prediction function $g$.
\For{epoch $= 1$ to $E$}
    \For{each batch $B$}
        \State \textbf{Update $\pi$:} minimize $\mathcal{L}_{rep}$:
        
        \State $\pi \gets \pi - \eta_{\pi} \nabla_{\pi} \big( \frac{1}{|B|} \sum_{(i, j) \in B} ||\mathbf{S}_{ij} - $
        \Statex $\pi(\phi(x_i), \phi(x_j))||_2^2 \big)$
        \State \textbf{Update $\phi$ and $g$:} minimize $\mathcal{L}_{pre}- \gamma \mathcal{L}_{rep}$:
        
        \State $\phi \gets \phi - \eta \nabla_{\phi} (-\frac{1}{|B|} \sum_{(i, j) \in B} \sum_{c=1}^{C} Y_{ij}^c $
        \Statex $\log ( g_c(\phi(x_i), \phi(x_j),  \mathbf{S}_{ij}) )- \gamma \mathcal{L}_{rep} )$
        \State $g \gets g - \eta \nabla_{g} ( -\frac{1}{|B|} \sum_{(i, j) \in B} \sum_{c=1}^{C} Y_{ij}^c $
        \Statex $\log ( g_c(\phi(x_i), \phi(x_j),\mathbf{S}_{ij})))$
    \EndFor
\EndFor
\State \textbf{Step 3: Edge Classification}
\State Predict edge classes $\hat{\mathbf{Y}} = g(\phi(x_i), \phi(x_j), \mathbf{S}_{ij})$, where $x_i$, $x_j$, and $\mathbf{S}_{ij}$ are from the test set.
\State \textbf{Return} Predicted edge classes $\hat{\mathbf{Y}}$.
\end{algorithmic}
\end{algorithm}

\subsection{Introduction of the Datasets}\label{app:datasets}
A brief summary of each dataset is provided below:
\begin{itemize}
    \item  \textbf{Bitcoin-alpha} \cite{derr2017signed}: A cryptocurrency transaction trust network with 3,784 nodes and 24,207 edges. Nodes represent users, edge features are text embeddings of user comments. Edge classes are ratings (how much one user trusts/distrusts another user), which range from -10 (total distrust) to +10 (total trust). The dataset can be used for binary or multi-class classification. In binary classification, trust (ratings 1 to 10) is the class 0, while distrust (ratings -10 to -1) is the class 1. For multi-class classification, weak trust, strong trust, weak distrust, and strong distrust are considered.
    \item \textbf{HSPPI}: A homo sapiens protein-protein interaction network based on the String database \cite{szklarczyk2023string} with 17,895 nodes and 1,008,006 edges. Nodes represent proteins with node features being protein sequences. Edge features are the co-expression. Edges classes are direct or indirect functional interactions between proteins.
    \item \textbf{NID}: From the NF-ToN-IoT dataset \cite{sarhan2021netflow}, representing a NetFlow-based IoT network with 109,802 nodes and 431,372 edges. Nodes are user IP addresses, and edge features consist of network flow statistics (flow duration, bytes, packets, etc.). Edges classes are interactions between users, which include nine classes representing various network activities.
    \item \textbf{Epinions} \cite{guha2004propagation}: A product review website with 81,350 nodes and 763,538 edges. Nodes are users, edges classes are trust or distrust, and edge features are created by concatenating the text embeddings of the source and target node reviews.
    \item \textbf{Reddit} \cite{kumar2018community}: A hyperlink network with communities as nodes and posts between them as edges. It has 67,180 nodes and 901,056 edges. Edge features are text property vectors between the communities. Edge classes describe the sentiments of posts (positive/neutral or negative). 
    \item \textbf{MAG} \cite{sinha2015overview}: A citation network where nodes represent scholars and edges represent co-authored papers. It has 40,000 nodes and 141,725 edges. Edge features and classes are abstract and the field of study, respectively. 
\end{itemize}

\subsection{Implementation Details of Main Results}
\label{appendix:Implementation Details of Main Results}
\textbf{Combining CECF with TopoEdge}\quad When combining our framework with TopoEdge, we retain the core architecture of TopoEdge and simply augment its loss function by adding our adversarial balanced counterfactual representation term \(-\gamma \mathcal{L}_{rep}\). Specifically, the combined loss function is defined as:
\begin{equation}
    \mathcal{L}_{\text{combined}} = \mathcal{L}_{\text{TopoEdge}} - \gamma \mathcal{L}_{rep},
\end{equation}
where \(\mathcal{L}_{\text{TopoEdge}}\) is the original loss function of TopoEdge, and \(\gamma\) is a hyperparameter that balances the contribution of our counterfactual representation term. Therefore, during training, we update \(\pi\) using gradient ascent to maximize \(\mathcal{L}_{rep}\), while the rest of the model parameters are updated using gradient descent to minimize \(\mathcal{L}_{\text{combined}}\).

\noindent \textbf{Parameter Settings for Baselines}\quad  
For all baseline methods (GCN, GAT, ChebNet, and TopoEdge), we adopt the hyperparameters provided in the original TopoEdge paper \cite{cheng2024edge}. This ensures a fair comparison and consistency with the established benchmark. 

\noindent\textbf{Parameter Settings for CECF}\quad   
In our framework, to have a fair comparison, we adopt the same settings for the hidden dimension and most of the hyperparameters as \cite{cheng2024edge}. Due to the different datasets having different numbers of edge features and different dimensions of node embeddings, we also detail some specific dimensional settings in Table~\ref{tab:layer_dim}. 
We only adjusted the learning rates for \(g\), \(\phi\), and \(\pi\), as well as the balancing coefficient \(\gamma\) \footnote{An empirical choice for $\gamma$ can be found in Appendix \ref{appendix:More experience}.}, which are tuned on the validation set. The functions \(g\) and \(\phi\) share a same learning rate for simplicity and efficient updates, while \(\pi\) uses another learning rate.    
These specific parameter values can be found in our publicly released code after the review.

\begin{table}[htbp]
  \centering
  \caption{The dimensions of our CECF on the different datasets. The Edge means the numbers of edge features, the Node means the dimensions of node embedding, the Cross-attention means the input dimension of the cross-attention, and the MLP means the output dimension of the MLP. 2/4 is for Binary and Multi-class tasks, respectively.}
     \setlength{\tabcolsep}{1mm} 
     \small
    \begin{tabular}{c|cccc}
    \toprule
          Dataset & Edge   & Node   & Cross-attention  & MLP  \\
    \midrule
          Bitcoin-alpha & 384   & 64   & 384+2$\times$64 & 2/4 \\
    \midrule
          PPI & 768   & 128   & 768+2$\times$128 & 2 \\
    \midrule
          Epinions & 768   & 256   & 768+2$\times$256 & 2 \\
    \midrule
          Reddit & 86   & 128   & 86+2$\times$128 & 2 \\
    \midrule
          NID & 8   & 256   & 384+2$\times$256 & 9 \\
    \midrule
          MAG & 256   & 128   & 256+2$\times$128 & 10 \\
    \bottomrule
    \end{tabular}
  \label{tab:layer_dim}
\end{table}%

\subsection{Implementation Details of Further Experiments}
\label{appendix:Implementation Details of Further Experiences}
In this subsection, we provide additional implementation details for the experiments conducted in Section \ref{sec:further_analysis}. 

\noindent\textbf{Canonical Correlation Analysis (CCA)}\quad  
CCA is a multivariate statistical method used to quantify the relationship between two sets of variables. In our case, we apply CCA to measure the interplay between node representations (\(z_i, z_j\)) and edge features (\(S_{ij}\)). For each backbone model (with and without CECF), we compute the canonical correlation coefficients on the test set, which represent the strength of the linear relationship between the two sets of variables.

\noindent\textbf{Shapley Value Calculations}\quad  
Shapley values are used to quantify the contribution of individual features to the model's predictions. Due to the computational complexity of calculating Shapley values for all samples, we uniformly sample 1,000 instances from the test set of each dataset for analysis. For each sampled instance, we compute the Shapley value for every feature, reflecting its contribution to the final prediction. To summarize the importance of node representations and edge features, we group the Shapley values into two categories: (1) node representations (\(z_i, z_j\)) and (2) edge features (\(S_{ij}\)). We then compute the average absolute Shapley value for each category, providing a clear measure of their relative importance in the model's decision-making process.

\label{appendix:dim_reduction}

\section{Theoretical Justification: $\ell_2$-Norm for High-Dimensional Balancing}
\label{app:theory}

In this section, we provide the theoretical justification for using the $\ell_2$-norm loss to achieve adversarial balancing in high-dimensional space.

\paragraph{Variational Equivalence.}
Following the variational inference framework \cite{kazemi2024adversarially} and recalling Equation (\ref{eq1:begin}) and (\ref{eq3}), our objective is to minimize the distributional discrepancy between the true edge features and the learned representation. While previous works focused on one-dimensional settings, the principle holds generally. In our high-dimensional context, by adopting a multivariate Gaussian variational approximation $q_{\pi}(T|Z) = \mathcal{N}(\pi(Z), \sigma^2 \mathbf{I})$, the maximization of the log-likelihood is equivalent to minimizing the Mean Squared Error (Equation (\ref{eq5:important})):
\begin{equation}
    \mathcal{L}_{rep} = \mathbb{E} \left[ \|\mathbf{S}_{ij} - \pi(Z)\|_2^2 \right].
\end{equation}
This confirms that the $\ell_2$ loss is the theoretically correct objective function derived from the causal objective.

\paragraph{Concentration Guarantee.}
We further justify that this $\ell_2$ objective is sufficient to bound the distributional deviation in high-dimensional spaces (dimension $d \gg 1$). We invoke the \textbf{Hanson-Wright Inequality} \cite{vershynin2018high}, a central result in high-dimensional probability.
Let $\mathbf{E} = \mathbf{S}_{ij} - \pi(Z)$ be the error vector with independent sub-Gaussian components. The inequality states that the Euclidean norm $\|\mathbf{E}\|_2$ concentrates sharply around its expectation:
\begin{equation}
    P\left( | \|\mathbf{E}\|_2 - \mathbb{E}[\|\mathbf{E}\|_2] | > t \right) \le 2 \exp\left( -c \min\left(\frac{t^2}{K^4}, \frac{t}{K^2}\right) \right),
\end{equation}
where $t > 0$ is the deviation threshold, $c$ is a constant, and $K$ is the sub-Gaussian norm.

In high-dimensional settings, the expected norm scales with the square root of the dimension (i.e., $\mathbb{E}[\|\mathbf{E}\|_2] \propto \sqrt{d}$). Consequently, for a fixed relative deviation fraction $\epsilon$ (i.e., setting $t = \epsilon \sqrt{d}$), the probability bound decays as $\exp(-C d)$.
This implies that constraining the $\ell_2$ norm via Equation (\ref{eq5:important}) imposes a geometrically stricter constraint as $d$ increases. The ``volume" of the solution space allowing for significant distributional deviations shrinks exponentially with $d$. Thus, minimizing the expected error ensures, with high probability, that the generated treatment distribution aligns uniformly with the true distribution.

\section{More Experiments}
\label{appendix:More experience}

\begin{table*}[htbp]
  \centering
  \caption{Performance (BACC / Macro-F1) of CECF (GCN Backbone) with Varying $\gamma$ Values. Best performance for each dataset-metric combination is highlighted in \textbf{bold}.}
  \label{tab:gamma_sensitivity_results}
  \small
  \begin{tabular}{@{}llccccc@{}}
    \toprule
    Dataset         & Metric & $\gamma=0$    & $\gamma=1 \times 10^{-4}$ & $\gamma=1 \times 10^{-3}$ & $\gamma=1 \times 10^{-2}$ & $\gamma=1 \times 10^{-1}$ \\
    \midrule
    \textbf{Bitcoin-alpha} & BACC   & 0.851         & 0.848                     & \textbf{0.866}            & 0.865                     & 0.843                     \\
                      & Macro F1     & 0.866         & 0.864                     & \textbf{0.867}            & 0.861                     & 0.864                     \\
    \midrule
    \textbf{Reddit}        & BACC   & 0.500         & 0.687                     & \textbf{0.772}            & 0.698                     & 0.498                     \\
                      & Macro F1     & 0.481         & 0.685                     & \textbf{0.776}            & 0.692                     & 0.476                     \\
    \midrule
    \textbf{MAG}           & BACC   & 0.574         & 0.671                     & \textbf{0.694}            & 0.180                     & 0.101                     \\
                      & Macro F1     & 0.584         & 0.676                     & \textbf{0.733}            & 0.169                     & 0.081                     \\
    \bottomrule
  \end{tabular}
\end{table*}

\subsection{The Ablation of $\mathcal{L}_{rep}$ } \label{appendix:The Ablation of loss}

\noindent\textbf{CECF without $\mathcal{L}_{rep}$}\quad  
In this implementation, we remove the causal inference learning component from the CECF framework, retaining only the original GNN encoder $\phi$ and the cross-attention layer as the prediction function $g$. Specifically, the node representations are directly obtained from the encoder $\phi$. Then, with the edge features $S_{ij}$ passed through an MLP, they are all fed into the function $g$ to predict the final edge class and trained using the loss $\mathcal{L}_{pre}$. The training parameters are consistent with those used in the main experiments.

\begin{table}[htbp]
    \renewcommand{\arraystretch}{0.8}
  \centering
  \caption{Edge Classification Performance on CECF without $\mathcal{L}_{rep}$ with different backbones.}
  \small
    \begin{tabular}{cc|ccc}
    \toprule
          & Backbone & GCN   & GAT   & ChebNet \\
    \midrule
    Dataset & Metrics & \multicolumn{3}{c}{Binary classification} \\
    \midrule
    \multirow{2}[2]{*}{Bitcoin-alpha} & BACC  & 0.833 & 0.621 & 0.641 \\
          & Macro F1 & 0.857 & 0.632 & 0.669 \\
    \midrule
    \multirow{2}[2]{*}{HSPPI} & BACC  & 0.531 & 0.501 & 0.590 \\
          & Macro F1 & 0.527 & 0.492 & 0.598 \\
    \midrule
    \multirow{2}[2]{*}{Epinions} & BACC  & 0.712 & 0.637 & 0.772 \\
          & Macro F1 & 0.742 & 0.675 & 0.788 \\
    \midrule
    \multirow{2}[2]{*}{Reddit} & BACC  & 0.559 & 0.542 & 0.524 \\
          & Macro F1 & 0.561 & 0.536 & 0.527 \\
    \midrule
    Dataset & Metrics & \multicolumn{3}{c}{Multi-class classification} \\
    \midrule
    \multirow{2}[2]{*}{Bitcoin-alpha} & BACC  & 0.502 & 0.526 & 0.381 \\
          & Macro F1 & 0.513 & 0.536 & 0.405 \\
    \midrule
    \multirow{2}[2]{*}{NID} & BACC  & 0.513 & 0.517 & 0.527 \\
          & Macro F1 & 0.502 & 0.504 & 0.514 \\
    \midrule
    \multirow{2}[2]{*}{MAG} & BACC  & 0.670 & 0.745 & 0.653 \\
          & Macro F1 & 0.698 & 0.724 & 0.678 \\
    \bottomrule
    \end{tabular}
  \label{tab:abla1}
\end{table}%

\textbf{$\mathcal{L}_{rep}$ is important for CECF as the causal modeling.} Comparing Table \ref{tab:abla1} with Table \ref{Binary} and \ref{Multi}, across the datasets, CECF without $\mathcal{L}_{rep}$ generally performs worse than the original CECF. For instance, in the Binary-class Bitcoin-alpha dataset, the BACC of GAT decreases by 0.236, and in the HSPPI dataset, it decreases by 0.10. This observation suggests that the performance improvement is not solely due to the replacement with the cross-attention layer, but rather the $\mathcal{L}_{rep}$ for the Adversarial Balanced Counterfactual Representation on Graphs, which enables more effective modeling of the causal relationships between nodes and edges. 

\subsection{The Ablation of $\gamma$ }
\label{The Ablation of gamma}
To investigate the impact of the hyperparameter $\gamma$ in Equation (\ref{eq5}), which balances the adversarial loss term ($\mathcal{L}_{\text{rep}}$) against the primary prediction loss ($\mathcal{L}_{\text{pre}}$), we conducted a sensitivity analysis. We empirically evaluated the performance of our CECF framework, using the GCN backbone, across a range of $\gamma$ values: $\{0, 1 \times 10^{-4}, 1 \times 10^{-3}, 1 \times 10^{-2}, 1 \times 10^{-1}\}$. This analysis was performed on three distinct datasets: Bitcoin-alpha (binary classification), Reddit (binary classification), and MAG (multi-class classification). The Balanced Accuracy (BACC) and Macro-F1 scores are presented in Table \ref{tab:gamma_sensitivity_results}.

The results in Table~\ref{tab:gamma_sensitivity_results} demonstrate a consistent trend across the evaluated datasets. As measured by both BACC and Macro-F1, the performance generally improves as $\gamma$ increases from 0, reaching an optimal or near-optimal level around $\gamma = 1 \times 10^{-3}$. Further increases in $\gamma$ typically lead to a decline in performance, which can be particularly sharp for larger values (e.g., $1 \times 10^{-2}$ or $0.1$ on MAG and Reddit). Based on this sensitivity analysis across diverse datasets, a value of $\gamma = 1 \times 10^{-3}$ consistently yields strong performance or is very close to the optimal. This suggests that $\gamma = 1 \times 10^{-3}$ can serve as a robust default value for practical applications of CECF.

Furthermore, to deep discuss this, when $\gamma$ is set to 0, effectively removing the adversarial learning component, the performance is notably lower than when a moderate $\gamma$ is used. This underscores the importance and effectiveness of the adversarial balanced counterfactual representation learning in our CECF framework. Conversely, if $\gamma$ is too large, the adversarial constraint ($\mathcal{L}_{\text{rep}}$) may become overly dominant. This can hinder the model's ability to learn the primary edge classification task effectively, leading to a significant drop in performance as the model prioritizes satisfying the adversarial objective over accurate prediction.

\begin{table*}[htbp]
\renewcommand{\arraystretch}{0.6}
  \centering
  \caption{Comparing HSIC on Binary-class Edge Classification Datasets (Mean Values). The best performance is highlighted in bold. }
  \setlength{\tabcolsep}{0.8mm} 
  \small
  \begin{tabular}{ccccccccccccc}
    \toprule
    Dataset & \multicolumn{3}{c}{Bitcoin-alpha} & \multicolumn{3}{c}{HSPPI} & \multicolumn{3}{c}{Epinions} & \multicolumn{3}{c}{Reddit} \\
    \midrule
    Model & BACC  & Macro F1 & Time (S) & BACC  & Macro F1 & Time (S) & BACC  & Macro F1 & Time (S) & BACC  & Macro F1 & Time (S) \\
    \midrule
    GCN   & 0.798 & 0.836 & 9.483 & 0.500 & 0.432 & 198.538 & \textbf{0.756} & 0.757 & 133.076 & 0.744 & 0.766 & 277.184 \\
    +CECF & \textbf{0.866} & \textbf{0.867} & 24.726 & \textbf{0.577} &\textbf{0.577} & 285.963 & 0.733 & 0.760 & 376.499 & \textbf{0.772} & \textbf{0.776} & 479.069 \\
    +HSIC & 0.797 & 0.842 & 14.950 & 0.500 & 0.432 & 3207.855 & 0.755 & \textbf{0.763} & 645.781 & 0.681 & 0.704 & 4746.592 \\
    \midrule
    GAT   & 0.826 & 0.846 & 12.076 & 0.500 & 0.433 & 201.142 & \textbf{0.726} & \textbf{0.751} & 158.308 & 0.539 & 0.536 & 218.781 \\
    +CECF & \textbf{0.857} & \textbf{0.865} & 23.392 & 0.511 & 0.488 & 292.048 & 0.657 & 0.693 & 390.158 & \textbf{0.777} & \textbf{0.776} & 504.268 \\
    +HSIC & 0.810 & 0.848 & 17.746 & \textbf{0.652} & \textbf{0.659} & 3227.247 & 0.664 & 0.699 & 585.350 & 0.691 & 0.722 & 5266.738 \\
    \midrule
    ChebNet & 0.819 & 0.849 & 11.989 & 0.500 & 0.432 & 190.256 & \textbf{0.762} & 0.778 & 114.461 & 0.751 & 0.772 & 343.553 \\
    +CECF & \textbf{0.850} & \textbf{0.866} & 23.433 & \textbf{0.640} & \textbf{0.640} & 273.911 & 0.758 & \textbf{0.784} & 340.686 & \textbf{0.795} & \textbf{0.784} & 367.216 \\
    +HSIC & 0.804 & 0.845 & 15.554 & 0.526 & 0.482 & 3222.709 & 0.751 & 0.779 & 514.014 & 0.732 & 0.724 & 2989.896 \\
    \bottomrule
  \end{tabular}
  \label{HSICBinary}
\end{table*}%

\begin{table*}[htbp]
\renewcommand{\arraystretch}{0.8}
  \centering
  \caption{Comparing HSIC on Multi-class Edge Classification Datasets (Mean Values). The best performance is highlighted in bold. }
     \setlength{\tabcolsep}{0.8mm} 
     \small
     \begin{tabular}{cccccccccc}
    \toprule
    Dataset & \multicolumn{3}{c}{Bitcoin-alpha} & \multicolumn{3}{c}{NID} & \multicolumn{3}{c}{MAG} \\
    \midrule
    Model & BACC  & Macro F1 & Time(S) & BACC  & Macro F1 & Time (S) & BACC  & Macro F1 & Time (S) \\
    \midrule
    GCN   & 0.514 & 0.531 & 15.209 & 0.523 & 0.510 & 148.578 & \textbf{0.797} & \textbf{0.834} & 95.890 \\
    +CECF & \textbf{0.537} & 0.534 & 24.936 & 0.523 & 0.507 & 199.373 & 0.694 & 0.733 & 158.244 \\
    +HSIC & 0.514 & \textbf{0.547} & 20.111 & \textbf{0.527} & \textbf{0.517} & 904.893 & 0.665 & 0.701 & 117.752 \\
    \midrule
    GAT   & 0.524 & 0.544 & 18.896 & 0.495 & 0.490 & 152.463 & \textbf{0.795} & \textbf{0.837} & 120.894 \\
    +CECF & \textbf{0.538} & 0.552 & 32.554 & \textbf{0.521} & \textbf{0.507} & 215.282 & 0.732 & 0.756 & 201.003 \\
    +HSIC & 0.523 & \textbf{0.563} & 21.249 & 0.474 & 0.475 & 900.246 & 0.659 & 0.714 & 131.315 \\
    \midrule
    ChebNet & \textbf{0.528} & 0.557 & 15.965 & 0.527 & 0.514 & 167.421 & \textbf{0.801} & \textbf{0.842} & 89.985 \\
    +CECF & 0.526 & \textbf{0.559} & 26.139 & 0.527 & 0.514 & 196.057 & 0.722 & 0.732 & 101.261 \\
    +HSIC & 0.500 & 0.547 & 20.497 & \textbf{0.566} & \textbf{0.537} & 857.795 & 0.634 & 0.676 & 104.331 \\
    \bottomrule
    \end{tabular}
  \label{HSICMulti}%
\end{table*}%

\subsection{The Ablation of Other Causal Frameworks}\label{app:Causal Frameworks}
Although our CECF is tailored for end-to-end high-dimensional graph learning with theoretical ground, we also discuss the feasibility of simpler statistical causal inference methods commonly used in low-dimensional settings. The core objective is to eliminate the confounding influence of node representations $Z$ on the edge features $\mathbf{S}$ (which serve as the treatment $T$ in our causal graph).
We consider two common approaches, Linear Residualization and Hilbert-Schmidt Independence Criterion (HSIC) Penalty. 

Linear Residualization operates by performing a linear regression of $\mathbf{S}$ on $Z$, predicting $\mathbf{S}$ via $Z$, and then using the residuals (the part of $\mathbf{S}$ uncorrelated with $Z$) for subsequent prediction. However, this method fundamentally relies on the assumption that both the covariates $Z$ and the treatment $\mathbf{S}$ are static, fixed features available prior to model training. 
In our graph scenario, this is unfeasible because the node representations $Z$ are latent vectors dynamically learned by the GNN model. We cannot perform a one-time regression on fixed data; rather, the representations evolve continuously. Consequently, unlike our adversarial learning framework which adapts dynamically, standard linear residualization cannot be seamlessly integrated into such an end-to-end training pipeline.

Alternatively, the Correlation or HSIC Penalty explicitly enforces independence by adding a regularization term to the loss function. While theoretically viable, we identify a critical limitations for its application in graph learning. Computing HSIC involves pairwise distance matrices, introducing a quadratic complexity $\mathcal{O}(B^2)$ (where $B$ is the batch size), which is computationally expensive for high-dimensional vectors.

Despite the limitation, to provide a comprehensive evaluation, we implemented the HSIC-based variant as a competitive baseline. In this variant, we replace the adversarial component of CECF with an HSIC regularization term. The HSIC measures the dependence between node embeddings $Z$ and edge features $\mathbf{S}$ by mapping them into Reproducing Kernel Hilbert Spaces (RKHS). The additional loss term is defined as:
\begin{equation}
    \mathcal{L}_{HSIC}(Z, \mathbf{S}) = \frac{1}{(B-1)^2} \text{Tr}(\mathbf{K} \mathbf{H} \mathbf{L} \mathbf{H}),
\end{equation}
where $B$ is the batch size, $\mathbf{K}, \mathbf{L} \in \mathbb{R}^{B \times B}$ are the Gram matrices computed using a Gaussian kernel for $Z$ and $\mathbf{S}$ respectively (i.e., $K_{ij} = k(z_i, z_j)$ and $L_{ij} = l(\mathbf{s}_i, \mathbf{s}_j)$), and $\mathbf{H} = \mathbf{I} - \frac{1}{B}\mathbf{1}\mathbf{1}^\top$ is the centering matrix. This term is added to the classification loss to penalize dependencies. As this loss term is combined with the original classification loss, we evaluated weight coefficients from the set $\{0.1, 0.5, 1\}$ and selected 0.5 as the optimal value based on validation set performance. The empirical results of this variant are presented in Table \ref{HSICBinary}  and \ref{HSICMulti}.

\textbf{The results demonstrate the superiority of our CECF framework over the HSIC-based variant in utilizing causal information.} First, CECF consistently outperforms HSIC across the vast majority of datasets and backbones. While the HSIC penalty yields marginal improvements over the original GNNs in some cases, it frequently struggles or even degrades performance in scenarios with strong node-edge interplay (e.g., Reddit and multi-class Bitcoin-alpha), whereas CECF achieves more gains. This indicates that our adversarial learning approach is more effective at modeling high-dimensional causal dependencies than a simple correlation penalty, which may fail to capture complex, non-linear confounding factors. Second, the results further validate our earlier conclusion regarding the specific applicability of causal frameworks. In the MAG dataset, which has weak node-edge interplay (low initial CCA), both causal methods (CECF and HSIC) exhibit a performance drop compared to the baselines. This consistent trend reinforces that explicit causal modeling is a specialized solution. It is highly effective for datasets where node features causally influence edge features, but may introduce unnecessary constraints when such causal links are absent.

\textbf{For computational efficiency, it also reveals the advantage of CECF.} While the HSIC penalty incurs manageable time costs on smaller datasets (e.g., Bitcoin-alpha), its computational burden becomes heavy as the dataset scale increases. Specifically, on larger datasets such as HSPPI and Reddit, the HSIC variant requires an excessive amount of training time, increasing by tens of times compared to the baseline (e.g., surging from $\sim$198s to over $\sim$3200s on HSPPI with GCN). This dramatic increase comes from the high computational complexity involved in calculating correlation matrices for high-dimensional vectors within batches. In contrast, CECF maintains a stable and modest overhead (keeping the total time within 2x that of the backbone), demonstrating that our CECF is much more practical for real-world graph learning tasks.

\subsection{Case Study}\label{app:case study}
To further understand how CECF improves performance, we conduct a case study analyzing specific instances where CECF corrects misclassifications on the GCN backbone in the test set of Bitcoin-alpha (binary). We investigate these ``flipped" cases by comparing their node and edge embedding similarities to the average embeddings of class 0 and class 1. This similarity determines a ``predictive tendency" for node and edge components. We then evaluate the accuracy of these individual tendencies and their combined effect (correct if either node or edge tendency is correct). This analysis aims to reveal: (1) if CECF enhances the class-discriminative quality of node and edge embeddings, and (2) if CECF better integrates node and edge information for correct final predictions. 

\begin{table}[htbp]
  \centering
  \renewcommand{\arraystretch}{0.8}
  \caption{Predictive Tendencies in Flipped Cases on Bitcoin-alpha (Binary). Counts of node, edge, and combined embedding tendencies aligning with the true class for 125 instances misclassified by GCN but corrected by GCN+CECF.}
  \small
    \begin{tabular}{cccc}
    \toprule
              & \makecell{Node \\ tendency} & \makecell{Edge \\ tendency} & \makecell{Combine \\ tendency} \\
    \midrule
    GCN   & 58    & 71    & 93 \\
    +CECF & 77    & 81    & 104 \\
    \bottomrule
    \end{tabular}
  \label{tab:case_study}
\end{table}%

The results in Table~\ref{tab:case_study}, analyzing 125 instances on Bitcoin-alpha misclassified by GCN but corrected by GCN+CECF, indicate that \textbf{CECF yields representations with more discriminative quality and learns better combine features for the predictions.} This is evident as CECF improves the alignment of individual embedding tendencies with the true class: node and edge tendency alignment increases from 58 to 77 cases and from 71 to 81 cases, respectively. This improvement in each representation quality underscores CECF's success in learning node and edge features. Furthermore, it worth noting that while GCN exhibited a correct combined tendency in 93 of these cases, its final prediction was still incorrect. In contrast, CECF not only achieves a higher correct combined tendencies (108) but also successfully corrects the classification for 125 instances. This demonstrates CECF's enhanced capability to foster a better integration between node and edge.

\subsection{The Detail Results of The Main Results}
\label{The Detail Results of The Main Results}
The detail results of Table \ref{Binary} and \ref{Multi} are shown in Table \ref{tab:The Detail Results of Binary} and \ref{tab:The Detail Results of Multi}.

\subsection{The Detail Results of The Further Experiments}\label{The Detail Results of The Further Experiences}
The detail results of Table \ref{tab:cca_shapley_main} are shown in Table \ref{tab:The Detail Results of cca_shapley_main}.

\section{Limitations and Future works}
\label{appendix: future work}
While we present a comprehensive exploration and detailed analysis of our Causal Edge Classification Framework (CECF), we recognize certain limitations that pave the way for promising future research directions. These include:

(1) \textbf{Interpretability and Explainability:} While causal graphs model relationships, the high-dimensional nature limits interpretability. Future work could leverage advanced visualization or post-hoc methods to enhance transparency and trust in the framework’s decisions.  

(2) \textbf{Integration with Domain Knowledge:} The current causal graph structure may not fully capture dataset-specific priors. Incorporating domain knowledge, especially in fields like bioinformatics or social network analysis, could refine causal graph construction and improve effectiveness.  

(3) \textbf{Scalability and Efficiency:} CECF’s adversarial learning introduces computational overhead. Future work could explore integrating confounding mitigation into the loss function to further reduce this burden.

\begin{sidewaystable}[htbp]
\renewcommand{\arraystretch}{0.6}
  \centering
  \caption{Binary-class Edge Classification Performance on Various Datasets (Mean $\pm$ Std. Dev.). The best performance is highlighted in bold. +CECF(TopoEdge) indicates that CECF is applied with the TopoEdge framework.}
    \resizebox{\textwidth}{!}{
    \setlength{\tabcolsep}{0.6mm} 
    \small
    \begin{tabular}{ccccccccccccc}
    \toprule
    Dataset & \multicolumn{3}{c}{Bitcoin-alpha} & \multicolumn{3}{c}{HSPPI} & \multicolumn{3}{c}{Epinions} & \multicolumn{3}{c}{Reddit} \\
    \midrule
    Model & BACC  & Macro F1 & Time (S) & BACC  & Macro F1 & Time (S) & BACC  & Macro F1 & Time (S) & BACC  & Macro F1 & Time (S) \\
    \midrule
    TER+AER(Poisson) & 0.822 $ \pm$ 0.095 & 0.833 $ \pm$ 0.052 & 23.997 $ \pm$ 0.113 & 0.537 $ \pm$ 0.028 & 0.508 $ \pm$ 0.052 & 123.106 $ \pm$ 0.948 & 0.734 $ \pm$ 0.006 & 0.776 $ \pm$ 0.003 & 62.362 $ \pm$ 0.982 & 0.721 $ \pm$ 0.002 & 0.763 $ \pm$ 0.002 & 217.828 $ \pm$ 1.059 \\
    TER+AER(Geometric) & 0.872 $ \pm$ 0.023 & 0.874 $ \pm$ 0.003 & 23.879 $ \pm$ 0.723 & 0.517 $ \pm$ 0.000 & 0.474 $ \pm$ 0.000 & 124.851 $ \pm$ 1.202 & 0.724 $ \pm$ 0.002 & 0.771 $ \pm$ 0.001 & 61.267 $ \pm$ 1.476 & 0.728 $ \pm$ 0.001 & 0.769 $ \pm$ 0.001 & 216.739 $ \pm$ 1.893 \\
    \midrule
    GCN   & 0.798 $ \pm$ 0.010 & 0.836 $ \pm$ 0.006 &   9.483 $\pm$ 0.506    & 0.500 $ \pm$ 0.000 & 0.432 $ \pm$ 0.000 &   198.538 $\pm$ 1.016    & \textbf{0.756 $ \pm$ 0.004} & 0.757 $ \pm$ 0.003 &   133.076 $\pm$ 19.523    & 0.744 $ \pm$ 0.004 & 0.766 $ \pm$ 0.001 & 277.184 $\pm$ 51.867 \\
    +TopoEdge & 0.862 $ \pm$ 0.020 & 0.859 $ \pm$ 0.002 &  12.553 $\pm$ 0.366     & 0.510 $ \pm$ 0.014 & 0.465 $ \pm$ 0.046 &   279.191 $\pm$ 0.744    & 0.747 $ \pm$ 0.006 & 0.764 $ \pm$ 0.000 &    179.555 $\pm$ 1.256   & 0.679 $ \pm$ 0.127 & 0.673 $ \pm$ 0.140 & 397.097 $\pm$ 109.736 \\
    +CECF & 0.866 $ \pm$ 0.006 & 0.867 $ \pm$ 0.003 &   24.726 $\pm$ 0.312    & 0.577 $ \pm$ 0.025 & 0.577 $ \pm$ 0.039 &   285.963 $\pm$ 1.200    & 0.733 $ \pm$ 0.009 & 0.760 $ \pm$ 0.003 &   376.499 $\pm$ 1.344    & \textbf{0.772 $ \pm$ 0.011} & \textbf{0.776 $ \pm$ 0.002} &  479.069 $\pm$ 17.403\\
    +CECF(Topoedge) & \textbf{0.872 $ \pm$ 0.008} & \textbf{0.871 $ \pm$ 0.002} &   26.801 $\pm$ 5.199    & \textbf{0.626 $ \pm$ 0.047} & \textbf{0.627 $ \pm$ 0.052} &   365.140 $\pm$ 0.650    & 0.748 $ \pm$ 0.008 & \textbf{0.765 $ \pm$ 0.001} &   398.645 $\pm$ 1.934    & 0.708 $ \pm$ 0.077 & 0.681 $ \pm$ 0.063 & 491.184 $\pm$ 114.738 \\
    \midrule
    GAT   & 0.826 $ \pm$ 0.007 & 0.846 $ \pm$ 0.006 &   12.076 $\pm$ 0.164    & 0.500 $ \pm$ 0.000 & 0.433 $ \pm$ 0.000 &   201.142 $\pm$ 1.780    & 0.726 $ \pm$ 0.010 & 0.751 $ \pm$ 0.002 &   158.308 $\pm$ 0.699    & 0.539 $ \pm$ 0.055 & 0.536 $ \pm$ 0.085 & 218.781 $\pm$ 50.260 \\
    +TopoEdge & 0.870 $ \pm$ 0.008 & 0.855 $ \pm$ 0.003 &   16.874 $\pm$ 4.771    & 0.500 $ \pm$ 0.000 & 0.433 $ \pm$ 0.001 &   277.797 $\pm$ 0.728    & 0.747 $ \pm$ 0.005 & 0.760 $ \pm$ 0.002 &    183.237 $\pm$ 1.345   & 0.500 $ \pm$ 0.000 & 0.475 $ \pm$ 0.000 & 245.266 $\pm$ 0.350 \\
    +CECF & 0.857 $ \pm$ 0.007 & 0.865 $ \pm$ 0.000 &   23.392 $\pm$ 5.407    & 0.511 $ \pm$ 0.010 & 0.488 $ \pm$ 0.040 &   292.048 $\pm$ 0.116    & 0.657 $ \pm$ 0.035 & 0.693 $ \pm$ 0.029 &   390.158 $\pm$ 4.354    & \textbf{0.777 $ \pm$ 0.012} & \textbf{0.776 $ \pm$ 0.004} & 504.268 $\pm$ 16.522 \\
    +CECF(Topoedge) & \textbf{0.872 $ \pm$ 0.010} & \textbf{0.870 $ \pm$ 0.006} &   23.262 $\pm$ 1.427    & \textbf{0.543 $ \pm$ 0.049} & \textbf{0.533 $ \pm$ 0.067} &   372.402 $\pm$ 1.933    & \textbf{0.750 $ \pm$ 0.009} & \textbf{0.763 $ \pm$ 0.001} &   413.414 $\pm$ 0.642    & 0.559 $ \pm$ 0.043 & 0.544 $ \pm$ 0.044 & 589.884 $\pm$ 69.786 \\
    \midrule
    ChebNet & 0.819 $ \pm$ 0.018 & 0.849 $ \pm$ 0.011 &   11.989 $\pm$ 2.334    & 0.500 $ \pm$ 0.000 & 0.432 $ \pm$ 0.000 &   190.256 $\pm$ 1.494    & 0.762 $ \pm$ 0.004 & 0.778 $ \pm$ 0.001 &   114.461 $\pm$ 5.046    & 0.751 $ \pm$ 0.006 & 0.772 $ \pm$ 0.006 & 343.553 $\pm$ 6.320 \\
    +TopoEdge & \textbf{0.882 $ \pm$ 0.011} & 0.871 $ \pm$ 0.001 &   21.697 $\pm$ 0.496    & 0.500 $ \pm$ 0.000 & 0.432 $ \pm$ 0.000 &   272.613 $\pm$ 1.193    & 0.760 $ \pm$ 0.005 & \textbf{0.784 $ \pm$ 0.001} &    171.193 $\pm$ 0.461   & 0.682 $ \pm$ 0.129 & 0.678 $ \pm$ 0.143 & 394.174 $\pm$ 108.398 \\
    +CECF & 0.850 $ \pm$ 0.010 & 0.866 $ \pm$ 0.002 &   23.433 $\pm$ 2.079    & 0.640 $ \pm$ 0.016 & 0.640 $ \pm$ 0.005 &   273.911 $\pm$ 0.693    & 0.758 $ \pm$ 0.006 & \textbf{0.784 $ \pm$ 0.002} &   340.686 $\pm$ 0.464    & \textbf{0.795 $ \pm$ 0.006} & \textbf{0.784 $ \pm$ 0.002} &  367.216 $\pm$ 32.092\\
    +CECF(Topoedge) & 0.877 $ \pm$ 0.003 & \textbf{0.874 $ \pm$ 0.005} &   26.318 $\pm$ 4.357    & \textbf{0.674 $ \pm$ 0.022} & \textbf{0.657 $ \pm$ 0.004} &   350.087 $\pm$ 1.783    & \textbf{0.764 $ \pm$ 0.006} & 0.721 $ \pm$ 0.007 &   346.676 $\pm$ 15.714    & 0.777 $ \pm$ 0.007 & 0.763 $ \pm$ 0.002 & 478.361 $\pm$ 5.693 \\
    \bottomrule
    \end{tabular}
    }
  \label{tab:The Detail Results of Binary}
\end{sidewaystable}%

\begin{sidewaystable}[htbp]
\renewcommand{\arraystretch}{0.6}
  \centering
  \caption{Multi-class Edge Classification Performance on Various Datasets (Mean $\pm$ Std. Dev.). The best performance is highlighted in bold. +CECF(TopoEdge) indicates that CECF is applied with the TopoEdge framework.}
     \setlength{\tabcolsep}{0.6mm} 
     \small
     \begin{tabular}{cccccccccc}
    \toprule
    Dataset & \multicolumn{3}{c}{Bitcoin-alpha} & \multicolumn{3}{c}{NID} & \multicolumn{3}{c}{MAG} \\
    \midrule
    Model & BACC  & Macro F1 & Time(S) & BACC  & Macro F1 & Time (S) & BACC  & Macro F1 & Time (S) \\
    \midrule
    TER+AER(Poisson) & 0.364 $ \pm$ 0.114 & 0.392 $ \pm$ 0.130 & 23.913$ \pm$ 0.543 & 0.324 $ \pm$ 0.010 & 0.329 $ \pm$ 0.011 & 62.122 $ \pm$ 2.601 & 0.643 $ \pm$ 0.037 & 0.704 $ \pm$ 0.034 & 44.780 $ \pm$ 0.992 \\
    TER+AER(Geometric) & 0.486 $ \pm$ 0.135 & 0.507 $ \pm$ 0.137 & 24.410 $ \pm$ 1.136 & 0.319 $ \pm$ 0.007 & 0.326 $ \pm$ 0.006 & 41.928 $ \pm$ 2.937 & 0.638 $ \pm$ 0.029 & 0.701 $ \pm$ 0.031 & 44.909 $ \pm$ 0.428 \\
    \midrule
    GCN   & 0.514 $ \pm$ 0.010 & 0.531 $ \pm$ 0.006 &   15.209 $\pm$ 0.761    & 0.523 $ \pm$ 0.001 & 0.510 $ \pm$ 0.002 &   148.578 $\pm$ 27.213    & 0.797 $ \pm$ 0.008 & 0.834 $ \pm$ 0.007 & 95.890 $\pm$ 0.179 \\
    +TopoEdge & 0.572 $ \pm$ 0.011 & 0.558 $ \pm$ 0.004 &   12.369 $\pm$ 0.080    & 0.627 $ \pm$ 0.036 & 0.491 $ \pm$ 0.025 &   200.487 $\pm$ 1.609    & \textbf{0.808 $ \pm$ 0.006} & \textbf{0.840 $ \pm$ 0.002} & 129.986 $\pm$ 0.167 \\
    +CECF & 0.537 $ \pm$ 0.007 & 0.534 $ \pm$ 0.008 &   24.936 $\pm$ 0.279    & 0.523 $ \pm$ 0.004 & 0.507 $ \pm$ 0.004 &   199.373 $\pm$ 2.342    & 0.694 $ \pm$ 0.021 & 0.733 $ \pm$ 0.015 & 158.244 $\pm$ 0.508 \\
    +CECF(Topoedge) & \textbf{0.577 $ \pm$ 0.018} & \textbf{0.564 $ \pm$ 0.006} &   29.585 $\pm$ 5.141    & \textbf{0.733 $ \pm$ 0.002} & \textbf{0.555 $ \pm$ 0.006} &   228.624 $\pm$ 0.817    & 0.767 $ \pm$ 0.005 & 0.813 $ \pm$ 0.001 & 189.877 $\pm$ 2.584 \\
    \midrule
    GAT   & 0.524 $ \pm$ 0.007 & 0.544 $ \pm$ 0.006 &   18.896 $\pm$ 0.793    & 0.495 $ \pm$ 0.009 & 0.490 $ \pm$ 0.007 &   152.463 $\pm$ 19.447    & 0.795 $ \pm$ 0.010 & 0.837 $ \pm$ 0.009 & 120.894 $\pm$ 0.186 \\
    +TopoEdge & 0.577 $ \pm$ 0.006 & \textbf{0.565 $ \pm$ 0.006} &   16.713 $\pm$ 0.598    & 0.558 $ \pm$ 0.044 & 0.484 $ \pm$ 0.009 &   206.151 $\pm$ 0.919    & \textbf{0.812 $ \pm$ 0.007} & \textbf{0.840 $ \pm$ 0.004} & 146.772 $\pm$ 0.688 \\
    +CECF & 0.538 $ \pm$ 0.014 & 0.552 $ \pm$ 0.013 &   32.554 $\pm$ 0.275    & 0.521 $ \pm$ 0.002 & 0.507 $ \pm$ 0.004 &   215.282 $\pm$ 2.372    & 0.732 $ \pm$ 0.018 & 0.756 $ \pm$ 0.011 & 201.003 $\pm$ 3.036 \\
    +CECF(Topoedge) & \textbf{0.579 $ \pm$ 0.003} & 0.548 $ \pm$ 0.007 &   39.390 $\pm$ 0.694    & \textbf{0.693 $ \pm$ 0.020} & \textbf{0.544 $ \pm$ 0.004} &   242.679 $\pm$ 0.589    & 0.790 $ \pm$ 0.005 & 0.810 $ \pm$ 0.004 & 237.027 $\pm$ 12.525 \\
    \midrule
    ChebNet & 0.528 $ \pm$ 0.009 & 0.557 $ \pm$ 0.008 &   15.965 $\pm$ 0.134    & 0.527 $ \pm$ 0.001 & 0.514 $ \pm$ 0.002 &   167.421 $\pm$ 1.572    & 0.801 $ \pm$ 0.006 & \textbf{0.842 $ \pm$ 0.007} & 89.985 $\pm$ 0.767 \\
    +TopoEdge & \textbf{0.588 $ \pm$ 0.009} & 0.578 $ \pm$ 0.006 &   19.352 $\pm$ 1.683    & 0.651 $ \pm$ 0.009 & 0.522 $ \pm$ 0.009 &   194.287 $\pm$ 1.240    & \textbf{0.802 $ \pm$ 0.001} & 0.833 $ \pm$ 0.004 & 119.371 $\pm$ 1.565 \\
    +CECF & 0.526 $ \pm$ 0.012 & 0.559 $ \pm$ 0.017 &   26.139 $\pm$ 0.459    & 0.527 $ \pm$ 0.000 & 0.514 $ \pm$ 0.000 &   196.057 $\pm$ 0.968    & 0.722 $ \pm$ 0.013 & 0.732 $ \pm$ 0.013 & 101.261 $\pm$ 11.101 \\
    +CECF(Topoedge) & 0.585 $ \pm$ 0.008 & \textbf{0.583 $ \pm$ 0.005} &   29.972 $\pm$ 3.037    & \textbf{0.683 $ \pm$ 0.013} & \textbf{0.561 $ \pm$ 0.003} &   219.061 $\pm$ 0.744    & 0.776 $ \pm$ 0.012 & 0.801 $ \pm$ 0.003 & 170.096 $\pm$ 5.747 \\
    \bottomrule
    \end{tabular}
  \label{tab:The Detail Results of Multi}%
\end{sidewaystable}%

\begin{table*}[htbp]
\renewcommand{\arraystretch}{0.6} 
  \centering
  
  \caption{Comparison of Canonical Correlation Analysis (CCA) and Shapley Values for baseline models and our final CECF model. Note: A value of 0.0000 does not strictly equal zero but is a result of rounding an extremely small actual value.}
  \label{tab:The Detail Results of cca_shapley_main}
  \small
\setlength{\tabcolsep}{1mm} 
\begin{tabular}{cc|cc|cc|cc}
    \toprule
          & Model & GCN   & +CECF & GAT   & +CECF & ChebNet & +CECF \\
    \midrule
    Dataset & Metrics & \multicolumn{6}{c}{Binary Classification} \\
    \midrule
    \multirow{3}[2]{*}{Bitcoin-alpha} & CCA          & 0.570  & 0.604  & 0.369  & 0.464  & 0.510  & 0.483  \\
          & Node Shapley & 0.0103 & 0.0480 & 0.0102 & 0.0306 & 0.0206 & 0.0733 \\
          & Edge Shapley & 0.0197 & 0.0254 & 0.0257 & 0.0089 & 0.0222 & 0.0187 \\
    \midrule
    \multirow{3}[2]{*}{HSPPI}         & CCA          & 0.272  & 0.421  & 0.218  & 0.290  & 0.351  & 0.310  \\
          & Node Shapley & 0.0000 & 0.7711 & 0.0012 & 2200.6812 & 0.0000 & 35230.5270 \\
          & Edge Shapley & 0.0000 & 0.0002 & 0.0000 & 0.0000 & 0.0000 & 0.0000 \\
    \midrule
    \multirow{3}[2]{*}{Epinions}      & CCA          & 0.803  & 0.814  & 0.480  & 0.759  & 0.860  & 0.838  \\
          & Node Shapley & 0.0013 & 0.0030 & 0.0003 & 0.0041 & 0.0021 & 0.0009 \\
          & Edge Shapley & 0.0060 & 0.0042 & 0.0325 & 0.0049 & 0.0000 & 0.0000 \\
    \midrule
    \multirow{3}[2]{*}{Reddit}        & CCA          & 0.368  & 0.371  & 0.410  & 0.446  & 0.384  & 0.326  \\
          & Node Shapley & 0.0000 & 0.0187 & 0.0002 & 0.0108 & 0.0013 & 0.0146 \\
          & Edge Shapley & 0.0410 & 0.0109 & 0.0000 & 0.0202 & 0.0436 & 0.0272 \\
    \midrule
    Dataset & Metrics & \multicolumn{6}{c}{Multi-class Classification} \\
    \midrule
    \multirow{3}[2]{*}{Bitcoin-alpha} & CCA          & 0.573  & 0.614  & 0.390  & 0.548  & 0.501  & 0.448  \\
          & Node Shapley & 0.0223 & 0.0423 & 0.0160 & 0.1487 & 0.0322 & 0.0844 \\
          & Edge Shapley & 0.0409 & 0.0052 & 0.0463 & 0.0715 & 0.0396 & 0.0267 \\
    \midrule
    \multirow{3}[2]{*}{NID}           & CCA          & 0.911  & 0.963  & 0.832  & 0.939  & 0.901  & 0.941  \\
          & Node Shapley & 0.0576 & 0.0925 & 0.0540 & 0.0443 & 0.0611 & 0.0404 \\
          & Edge Shapley & 1.0634 & 0.0265 & 0.6334 & 0.0608 & 0.8354 & 0.0301 \\
    \midrule
    \multirow{3}[2]{*}{MAG}           & CCA          & 0.156  & 0.324  & 0.130  & 0.324  & 0.078  & 0.165  \\
          & Node Shapley & 0.0006 & 0.0676 & 0.0002 & 0.0567 & 0.0005 & 0.0789 \\
          & Edge Shapley & 0.0612 & 0.0989 & 0.0422 & 0.0511 & 0.0550 & 0.1116 \\
    \bottomrule
\end{tabular}%
\end{table*}

\end{document}